\def\ARXIVVERSION{1}

\documentclass{article}
\usepackage{iclr2027_conference,times}

\ifdefined\ARXIVVERSION
\iclrfinalcopy
\fi

\usepackage{amsmath,amsfonts,bm}

\def\eqref#1{equation~\ref{#1}}

\def\1{\bm{1}}

\DeclareMathAlphabet{\mathsfit}{\encodingdefault}{\sfdefault}{m}{sl}
\SetMathAlphabet{\mathsfit}{bold}{\encodingdefault}{\sfdefault}{bx}{n}

\usepackage{amsmath,amssymb}
\usepackage{booktabs}
\usepackage{graphicx}
\usepackage{array}
\usepackage{tabularx}
\usepackage{xcolor}
\usepackage{colortbl}
\usepackage{microtype}
\usepackage{placeins}
\usepackage{float}
\usepackage{hyperref}
\usepackage{url}
\ifdefined\ARXIVVERSION
\hypersetup{
  hidelinks,
  pdftitle={Beyond Average Error through Oracle-Informed Stress Tests for Time-Series Forecasting},
  pdfauthor={Xu Lin, Runheng Zuo, Shengxuan Xu, Qitai Tan, Hongyu Lin, Xiao-Ping Zhang},
  pdfkeywords={time-series forecasting, forecasting benchmarks, stress testing, distribution shift, oracle attribution}
}
\else
\hypersetup{hidelinks}
\fi

\makeatletter
\def\paragraph{\@startsection{paragraph}{4}{\z@}{0.7ex plus
0.1ex minus .1ex}{-1em}{\normalfont\normalsize\usefont{T1}{ptm}{b}{n}}}
\makeatother
\definecolor{basegray}{HTML}{64748B}
\definecolor{shiftorange}{HTML}{E76F51}
\definecolor{bayesblue}{HTML}{8ECAE6}
\definecolor{excessblue}{HTML}{277DA1}
\definecolor{improvegreen}{HTML}{2A9D8F}
\newcommand{\Base}{\textsc{Base}}
\newcommand{\Shift}{\textsc{Shift}}
\newcommand{\MSEmu}{\operatorname{MSE}_{\mu}}
\newcommand{\SMSE}{\operatorname{SMSE}}

\title{Beyond Average Error through Oracle-Informed Stress Tests for Time-Series Forecasting}

\ifdefined\ARXIVVERSION
\author{
Xu Lin \quad Runheng Zuo \quad Shengxuan Xu \quad Qitai Tan \quad
Hongyu Lin \quad Xiao-Ping Zhang\\[3pt]
Tsinghua University
}
\else
\author{Anonymous authors}
\fi

\begin{document}
\raggedbottom

\maketitle
\ifdefined\ARXIVVERSION
\lhead{Preprint}
\fi

\begin{abstract}
Average squared error cannot reveal whether forecasting performance degrades because the future becomes
less predictable or because forecasts move farther from the conditional mean. We introduce paired,
mechanism-controlled stress tests that decompose changes in expected squared error at each lead time into
environmental risk and forecast--oracle distance, using an origin-conditioned predictive oracle unavailable
to the evaluated models. Three end-to-end controls have known attribution. Specifically, the null,
environmental-only, and information-gap controls verify that the pipeline assigns changes to the correct
component. We then apply the
benchmark to 24 deployable forecasters. Under frequent switching, 14 methods have higher realized MSE but
lower oracle distance; under outlier-variance feedback, 19 have higher MSE but lower
scale-standardized MSE. Short- and long-lead stress-response rankings have Spearman correlation 0.624,
revealing substantial horizon-dependent reordering. We then study multivariate relation shifts. Across six
models and three coupling severities, oracle distance accounts for only 0.7--3.9\% of the
decomposed expected-risk increase, and environmental-risk majority persists in an eight-channel system and
a matched-difficulty audit of Ring, Block, and Hub relations. Finally, prespecified contrasts on independent data-generating
process (DGP) realizations
show that several visually compelling discovery profiles, including trend accumulation and the
hypothesized switching reversal, do not replicate. The benchmark thus combines component-wise diagnosis
with a held-out stability audit. It complements real-data out-of-distribution evaluation, which measures
performance under realistic shifts when exact oracle attribution is unavailable.
\end{abstract}

\section{Introduction}

Suppose two environments have the same conditional mean but different innovation variance.  The same
forecast has higher expected squared error in the noisier environment although its distance from the
predictable signal is unchanged. Conversely, similar average error can conceal failure under one mechanism
or at long leads. Raw MSE is therefore essential operationally but insufficient for attribution because it does
not distinguish environmental uncertainty from oracle distance, which itself includes the
information restriction imposed on the model.

This distinction guides action. Stable oracle distance with rising raw risk points toward
uncertainty management; rising distance instead motivates changes to information, model, or adaptation.
Mechanism-by-lead reporting also prevents a short-lead or benign-condition success from determining a
stressed long-horizon choice through an average alone.

\begin{figure}[t]
\centering
\includegraphics[width=\textwidth]{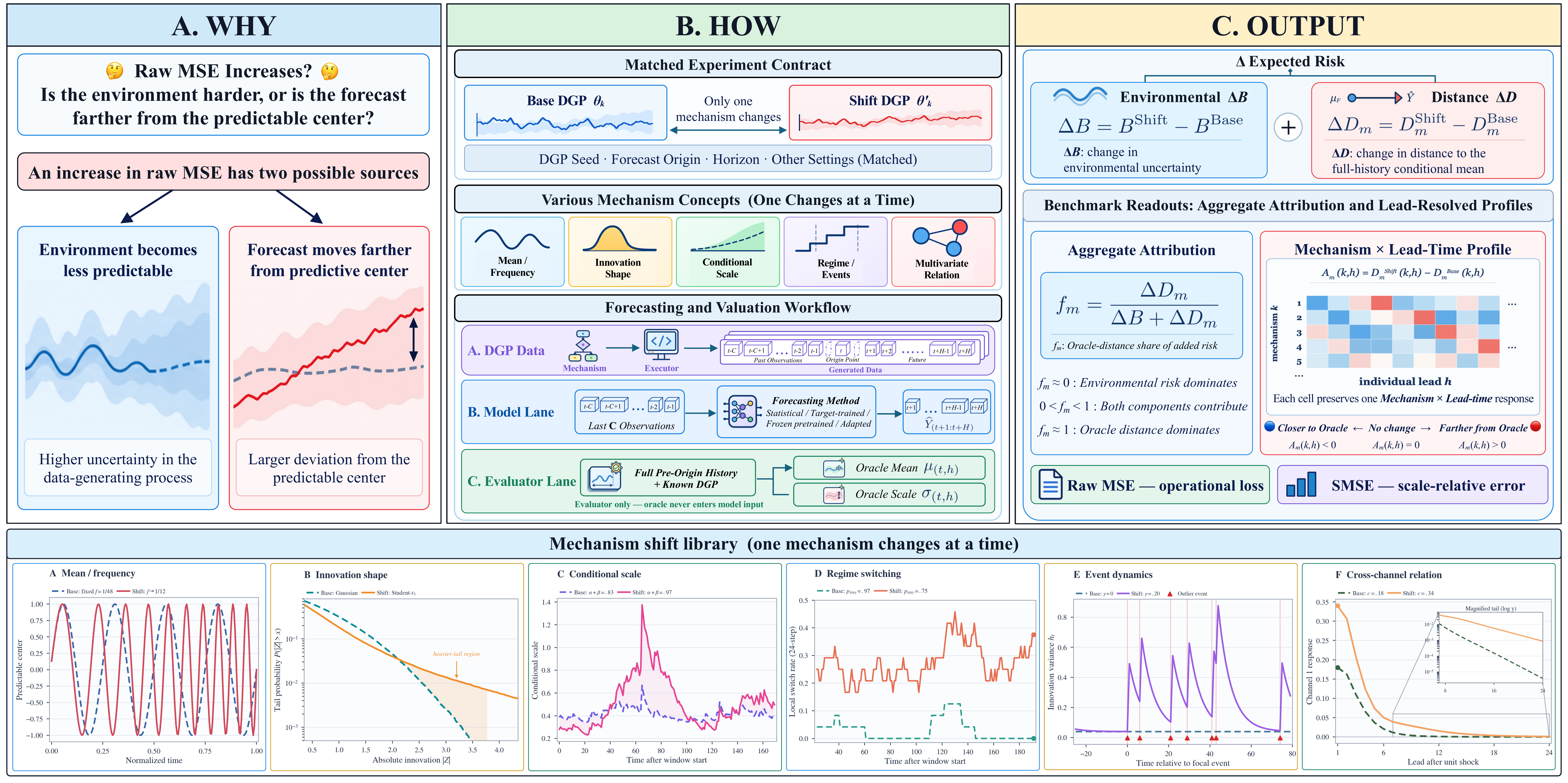}
\caption{\textbf{Overview of our oracle-informed, mechanism-controlled stress-test framework, which uses
matched pairs of \Base{} and \Shift{} environments across six mechanism families to separate environmental-risk changes from
changes in oracle distance and report aggregate and lead-resolved diagnostics.}}
\label{fig:overview}
\end{figure}

Real-data benchmarks broaden coverage, while controlled benchmarks isolate temporal or financial
mechanisms~\citep{godahewa2021monash,qiu2024tfb,aksu2024gifteval,tan2025syntsbench,sun2026finstressts}.
Neither supplies the complete combination used here, which consists of strict pairing between \Base{} and
\Shift{}, an origin-conditioned predictive oracle, and mechanism-by-individual-lead reporting
(Appendix Table~\ref{tab:prior-scope}). We
ask how much controlled forecast-risk degradation is environmental rather than oracle distance,
and how that attribution varies across mechanisms and leads. Figure~\ref{fig:overview} summarizes the paired
design. In each pair, one named mechanism changes, the known DGP supplies the evaluator-only predictive oracle,
and raw, oracle-distance, scale-relative, and lead-resolved outputs are reported together.

Across 24 deployable forecasters, the univariate atlas reveals structured diagnostic disagreements.
Specifically, frequent switching often raises realized MSE while reducing oracle distance, outlier-variance feedback
often raises MSE while reducing scale-relative error, and response rankings reorder substantially between
short and long leads. The multivariate analysis then provides the strongest replicated attribution result.
In a four-channel VAR, oracle distance accounts for only 0.7--3.9\% of the decomposed
expected-risk increase across six models and three coupling severities. Environmental-risk majority also
appears in an independently generated eight-channel system and matched-difficulty Ring, Block, and Hub
relations. Independent confirmation finally shows that several visually compelling univariate profiles do
not replicate, distinguishing exploratory structure from stable curve-level findings.

Our contributions are threefold.
\begin{itemize}
    \item a paired, lead-resolved measurement framework separating environmental risk from
    oracle distance while exposing model information restrictions;
    \item a validated benchmark with six controlled mechanism families, common forecasting interfaces,
    known-attribution controls, and held-out confirmation; and
    \item evidence from 24 deployable forecasters that diagnostic directions and rankings depend on
    mechanism and lead, while coupling degradation is predominantly environmental across the evaluated
    severities and structures.
\end{itemize}

\section{Problem formulation}
\label{sec:formulation}

\paragraph{Forecasting contract.}
At a forecast origin $t$, a model $m$ receives only the context
$X_t=(Y_{t-C},\ldots,Y_{t-1})$ and a requested horizon $H$.  In the multivariate extension,
$X_t\in\mathbb R^{C\times d}$.  It returns a point forecast
$\hat{\mathbf Y}_{m,t}=(\hat Y_{m,t},\ldots,\hat Y_{m,t+H-1})$ and optional quantiles. The forecast
origin $t$ denotes the first unobserved index; the context ends at $t-1$, and lead $h$ targets
$Y_{t+h-1}$. The model never receives DGP parameters, latent or realized futures, or oracle paths. Let
$\mathcal F_t^{\mathrm{obs}}=\sigma(Y_1,\ldots,Y_{t-1})$ denote the complete observed history and
$\mathcal F_t^{\mathrm{eval}}=\sigma(\mathcal F_t^{\mathrm{obs}},Z_{<t}^{\mathrm{DGP}})$ the evaluator
information set, where $Z_{<t}^{\mathrm{DGP}}$ contains only explicitly declared pre-origin generator
state. Finally, $\mathcal G_t^m=\sigma(X_t)\subseteq\mathcal F_t^{\mathrm{obs}}\subseteq
\mathcal F_t^{\mathrm{eval}}$ is the model-visible information; fitted parameters $W_m$ are fixed before
the test forecast. No future latent state, event, or observation enters either information set.

\paragraph{Risk targets and the information gap.}
For environment $e$, mechanism $k$, and lead time $h$, let
$\mu^{e,F}_{t,h}=\mathbb E[Y^e_{t+h-1}\mid\mathcal F_t^{\mathrm{eval}}]$,
$\mu^{e,G}_{m,t,h}=\mathbb E[Y^e_{t+h-1}\mid\mathcal G_t^m]$, and
$v^e_{t,h}=\operatorname{Var}(Y^e_{t+h-1}\mid\mathcal F_t^{\mathrm{eval}})$.
Here, $\mu^{e,F}_{t,h}$ is the evaluator-only predictive mean, $\mu^{e,G}_{m,t,h}$ uses model-visible
information, and $v^e_{t,h}$ is the remaining uncertainty under the declared evaluator information set.
The conditional identity is
\begin{equation}
\mathbb E[(Y^e_{t+h-1}-\hat Y^e_{m,t+h-1})^2\mid\mathcal F_t^{\mathrm{eval}}]
=\underbrace{v^e_{t,h}}_{\text{environmental risk}}
+\underbrace{(\hat Y^e_{m,t+h-1}-\mu^{e,F}_{t,h})^2}_{\text{instance-level forecast--oracle distance}}.
\label{eq:decomp}
\end{equation}
Suppressing indices, averaging the second term over evaluation histories gives
\begin{equation}
\mathbb E(\hat Y_m-\mu^F)^2
=\underbrace{\mathbb E(\mu^F-\mu_m^G)^2}_{\text{information restriction}}
+\underbrace{\mathbb E(\mu_m^G-\hat Y_m)^2}_{\text{model approximation and estimation}}.
\label{eq:information-gap}
\end{equation}
We denote the left-hand side by $D_m^e(k,h)$. Thus forecast--oracle distance contains both information
restriction and model approximation and estimation; it is not pure architecture error. Raw risk measures
operational loss, whereas $v^e_{t,h}$ measures evaluator-conditioned environmental uncertainty. Appendix
\ref{app:decomp-proof} gives assumptions and proof. The identity is specific to squared loss and its
conditional-mean Bayes act.

The formal DGPs provide analytic evaluator-conditioned means, so over $N$ matched instances we compute
\begin{equation}
\widehat D_m^e(k,h)=\frac{1}{N}\sum_{i=1}^N
(\hat Y^{e,(i)}_{m,t+h-1}-\mu^{e,F,(i)}_{t,h})^2.
\label{eq:oracle-distance}
\end{equation}
where $i$ indexes instances. Analytic means avoid finite-path bias; Monte Carlo futures estimate scale and
distributional quantities. Appendix~\ref{app:mc-audit} derives the simulation correction and audits
$S\in\{2048,4096,8192\}$.

Where analytic conditional variance is available, the evaluator can also form the controlled
expected-risk audit
\begin{equation}
R_{\mathrm{oracle},m}^{e}(t,h)=v^e_{t,h}+
(\hat Y^e_{m,t+h-1}-\mu^{e,F}_{t,h})^2,
\label{eq:oracle-risk}
\end{equation}
This oracle-moment audit complements realized MSE on the sampled future.

For a paired contrast between \Base{} and \Shift{}, let $B^e$ and $D_m^e$ denote the environmental-risk and
oracle-distance terms after the same prespecified aggregation over instances, channels, origins, and
leads.  We define
\begin{equation}
\Delta B=B^{\Shift}-B^{\Base},\qquad
\Delta D_m=D_m^{\Shift}-D_m^{\Base},\qquad
\Delta R_{\mathrm{oracle},m}=\Delta B+\Delta D_m.
\label{eq:delta-decomp}
\end{equation}
When $\Delta R_{\mathrm{oracle},m}>0$, the added-distance fraction is summarized by
\begin{equation}
f_m=\frac{\Delta D_m}{\Delta B+\Delta D_m}.
\label{eq:added-fraction}
\end{equation}
We report both components and interpret $f_m$ only for positive total degradation. The prespecified
$f_m\leq0.10$ diagnostic margin assigns at least 90\% of the increase to environmental risk; continuous
estimates and simultaneous bounds permit other tolerances.

\paragraph{Paired stress profile.}
The oracle-distance effect of a controlled stress is
\begin{equation}
A_m(k,h)=\widehat D_m^{\Shift}(k,h)-\widehat D_m^{\Base}(k,h).
\label{eq:profile}
\end{equation}
$A_m(k,h)>0$ means the forecast moves farther from the shifted environment's conditional mean; a negative
value does not imply lower operational risk because environmental risk may rise. The stress profile
$\mathcal P_m=\{A_m(k,h)\}_{k,h}$ retains mechanism and individual lead instead of only the aggregated
$\Delta D_m$. We additionally report SMSE, realized squared error divided pointwise by oracle predictive
variance; its exact definition and sensitivity analyses appear in Appendix~\ref{app:smse-sensitivity}.

\section{Paired benchmark design}
\label{sec:benchmark}

\subsection{Controlled data for mechanism attribution}

Observed data rarely provide a single-mechanism counterfactual or known conditional moments. Controlled
generators provide both for Eq.~\ref{eq:decomp}; descriptive external checks only probe sensitivity beyond
these formal DGPs.

Each univariate series has the common form
\begin{equation}
Y_t=\mu_t(\theta_\mu)+\sigma_t(\theta_\sigma,\mathcal F_t^{\mathrm{eval}})Z_t+J_t,
\label{eq:dgp}
\end{equation}
with predictable center, conditional scale, standardized innovation, and optional events. A paired
\Base{} and \Shift{} comparison changes one named parameter while matching DGP seed, origin, horizon, and all
non-target settings; the observations themselves come from their respective environments.

The six families cover mean and frequency, innovation shape, conditional scale, regime switching, event
dynamics, and multivariate cross-channel relation. Representative contrasts appear in Figure
\ref{fig:overview}; complete equations, parameters, and controls are in Appendix~\ref{app:dgp}.

\subsection{Forecast-origin predictive oracle}

The evaluator conditions on $\mathcal F_t^{\mathrm{eval}}$, while the model uses $\mathcal G_t^m$.
Markov state probabilities and GARCH scale are reconstructed strictly from pre-origin observations.
Persistent-level events, stochastic-volatility scale, and outlier-variance feedback additionally use their
recorded pre-origin generator state $Z_{<t}^{\mathrm{DGP}}$; future state, future events, and realized
future observations remain excluded. Means are analytic; sampled futures provide scale and distributional
scores. Oracle quantities remain evaluator-only, with path-count sensitivity in
Appendix~\ref{app:mc-audit}. Consequently, forecast--oracle distance includes the model's information
restriction relative to this evaluator oracle.

\subsection{Benchmark interfaces and recorded artifacts}

The generator returns data plus evaluator metadata; the forecaster maps the last $C=512$ observations to
$H$ predictions; and the evaluator combines forecasts, realized futures, oracle quantities, and pairing
keys to compute the registered metrics. This separation prevents oracle leakage into model input.

\section{Evaluation protocol}
\label{sec:protocol}

\paragraph{Common grid.}
The main univariate evaluation uses 10 DGP seeds, two forecast origins separated by 192 time steps,
$C=512$, and $H\in\{1,24,96,192\}$. Five univariate contrasts form a common 38-cell grid for all 25
reported entries; multivariate relation shifts are separate. DGP seed is the independent unit ($n=10$)
after averaging origins and model initializations within seed. This is the \emph{discovery grid}.

\paragraph{What each evaluation protocol measures.}
A shift may alter fitting data, fixed-model context, or post-shift updates, so protocols remain separate.
Univariate target-trained models fit complete \Base{} and \Shift{} paths separately and measure a learning
procedure response. Multivariate \Base{} and \Shift{} pairs share pre-change data and weights, measuring immediate
OOD and later passive contextual response without updates. Frozen TSFMs keep one checkpoint; matched
adaptation permits budget-controlled environment-specific updates. Appendix Table~\ref{tab:protocols}
states the complete relations between training and evaluation, together with the exact origins.

The grid contains 15 target-trained architectures, five frozen TSFMs, four statistical forecasters, and
one evaluator-only oracle reference. Thus, 24 entries are deployable. It spans linear, recurrent, multiscale,
Transformer, and pretrained designs~\citep{zeng2023dlinear,nie2023patchtst,liu2024itransformer,
ansari2024chronos,das2024timesfm,woo2024moirai}; specifications are in Appendix~\ref{app:model-specs}.

\paragraph{Discovery and held-out confirmation.}
Before held-out execution, five discovery-derived scalar contrasts fixed their directions, lead blocks,
model panels, tests, and multiplicity families. Three disjoint ten-seed batches, labelled A, B, and C,
assess replication;
the prespecified All-30 analysis adds precision. H1 compares the late trend-versus-switching interaction
of Chronos-2 and TimesFM; H2 tests TSMixer's trend late-minus-early contrast; H3 and H4 test early and late
switching effects for the fixed AR, Chronos-2, TimesFM, and TSMixer panel; and H5 compares the late
trend-versus-switching interaction of SegRNN and Non-stationary Transformer (NST). Full formulas and lock
records are in Appendix~\ref{app:lock}.

\paragraph{Metrics and inference.}
Primary views are realized raw MSE, oracle distance, and SMSE. DGP seed remains the independent
unit. Discovery profiles are descriptive; confirmation uses prespecified scalar contrasts, effect
intervals, direction-matched tests, and within-family Holm correction~\citep{holm1979}. Full statistical and omnibus
audits appear in Appendix Table~\ref{tab:confirmation} and Appendix~\ref{app:omnibus}.

\section{Main results}
\label{sec:results}

\subsection{Validation on known-attribution controls}

The end-to-end pipeline recovers the intended attribution when the correct answer is known in advance.
An identical-data control returns zero change in every paired component for all 400 profiles spanning
models and seeds.
Increasing innovation scale while preserving the predictable center assigns all but 0.20\% of the
expected-risk increase to environmental risk, whereas deleting the latest AR(1) input leaves
$\Delta B=0$, produces a positive $\Delta D=.000315$, and therefore has $f=100\%$. These controls validate the route from generation
and fitting through prediction, pairing, oracle construction, and attribution; complete pass rules,
precision, and intervals appear in Appendix Table~\ref{tab:stage4-controls}.
Here, calibrated means validated against end-to-end controls with known attribution, not probabilistic
forecast calibration.

\subsection{Mechanism- and lead-resolved findings across 24 forecasters}

\begin{figure}[t]
\centering
\includegraphics[width=\textwidth]{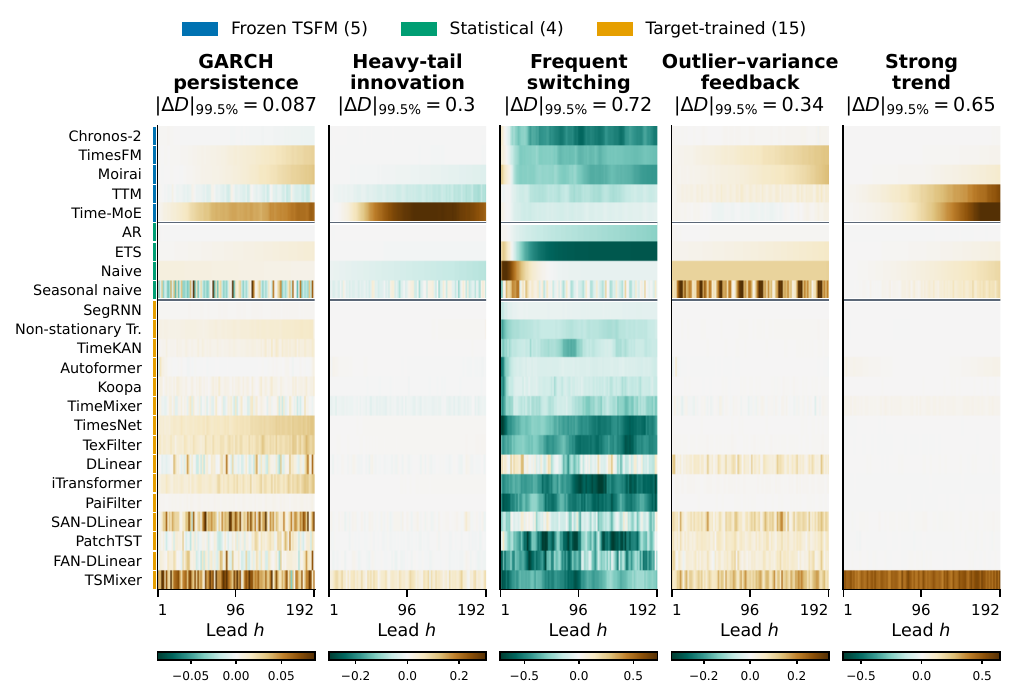}
\caption{\textbf{The complete discovery atlas retains model, mechanism, and individual lead.} Rows are
the 24 deployable forecasters, grouped by protocol; columns within each panel are the 192 individual leads
at $H=192$. Brown indicates increased oracle distance under \Shift{}, green indicates a
reduction. Each mechanism uses an explicitly labelled robust
symmetric range, so color intensity is compared within rather than across panels. These ten-seed profiles
support descriptive comparison and hypothesis generation; the stability of selected patterns is evaluated
on independent batches below.}
\label{fig:response-atlas}
\end{figure}

Figure~\ref{fig:response-atlas} shows how stress responses vary jointly with mechanism, forecast lead,
and forecasting method. The discovery atlas contains 24 deployable forecasters, consisting of 15
target-trained methods, five frozen TSFMs, and four statistical forecasters. It retains all 192 leads
for five representative univariate mechanism contrasts. It is a descriptive map for comparison and
hypothesis generation rather than a collection of confirmed model signatures. Complete aggregate metrics
for the 24 methods and the evaluator-only oracle reference are reported in Appendix
Table~\ref{tab:all-model-aggregate}.

\begin{figure*}[t]
\centering
\includegraphics[width=\textwidth]{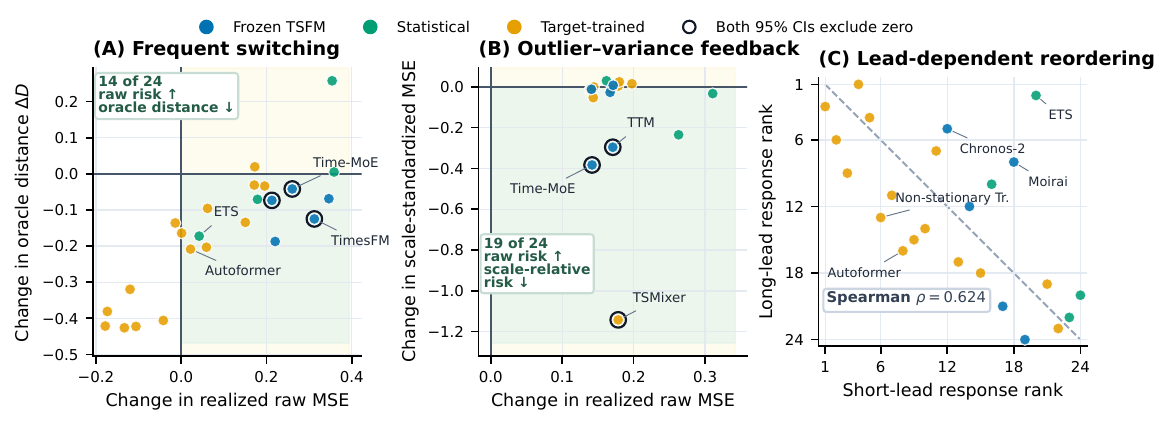}
\caption{\textbf{Aggregate error conceals diagnostic disagreement and lead-dependent ordering.}
(A) compares the change in realized raw MSE with the change in oracle distance under frequent
switching. (B) compares the change in realized raw MSE with the change in SMSE under
outlier-variance feedback. Shaded lower-right
quadrants mark increased raw loss accompanied by a lower diagnostic quantity; black rings mark methods
whose two unadjusted seed-bootstrap intervals exclude zero. (C) compares descriptive stress-response
ranks for short leads 1 through 24 and long leads 97 through 192; rank 1 denotes the lower weighted response score and the dashed
diagonal denotes no reordering. Colors identify forecasting protocol.}
\label{fig:diagnostic-disagreement}
\end{figure*}

Figure~\ref{fig:diagnostic-disagreement} shows that different diagnostic quantities and lead ranges can
produce materially different views of the same forecasts. Under frequent switching, 14 of 24 methods have higher
realized raw MSE but lower oracle distance: 7/15 target-trained, 5/5 frozen TSFMs, and 2/4 statistical
forecasters. Under outlier-variance feedback, 19 of 24 have higher realized raw MSE but lower SMSE:
12/15, 4/5, and 3/4, respectively. These pooled counts summarize the breadth of the diagnostic
disagreement; they do not define a common robustness estimand across training protocols. Short- and
long-lead stress-response rankings correlate only
$\rho=0.624$, with two methods moving by ten or more positions. Thus changes in the predictive center
can separate raw risk from oracle distance, changes in predictive scale can separate absolute
from relative error, and horizon aggregation can conceal model reordering.

The univariate atlas shows that aggregate forecast error can conceal both mechanism-specific and
horizon-dependent responses. We next apply the attribution framework to multivariate relation shifts,
where a change in cross-channel dependence can alter both environmental uncertainty and a forecaster's
distance from the predictive center.

\subsection{Environmental attribution under multivariate relation shifts}

\begin{figure}[t]
\centering
\includegraphics[width=\linewidth]{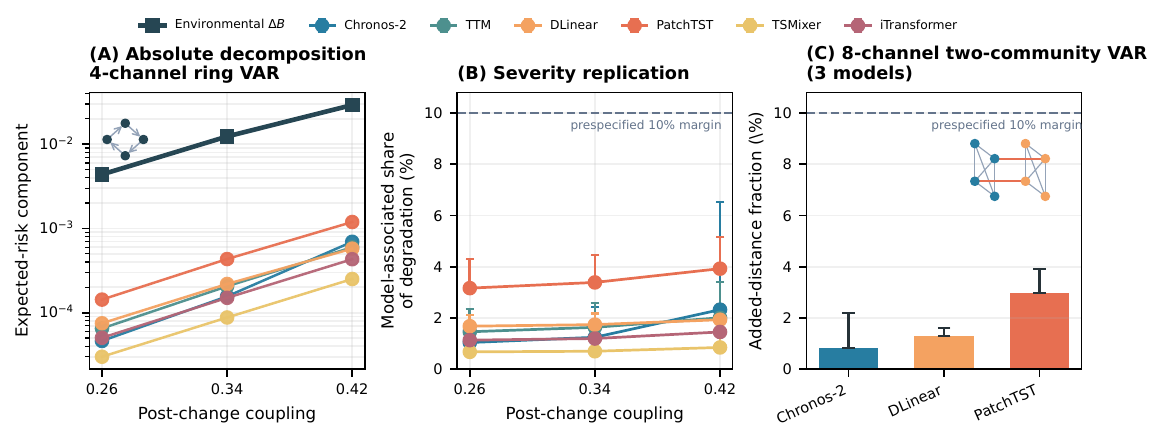}
\caption{\textbf{Coupling degradation is dominated by environmental risk across severity and structure.}
(A) Environmental $\Delta B$ (black squares) and model-specific $\Delta D$ (colored circles) on a log
scale as four-channel post-change coupling increases from the common 0.18 Base. (B) Added-distance
fractions and simultaneous one-sided 95\% upper bounds over 20 new seeds. (C) The same attribution in a
second eight-channel VAR for three prespecified models. The dashed line marks the prespecified 10\%
practical-significance criterion.}
\label{fig:attribution}
\end{figure}

Figure~\ref{fig:attribution} shows that environmental risk dominates coupling-shift degradation across
severity and the two displayed multivariate systems. In the
four-channel Ring system, a 20-seed replication fixes Base coupling at 0.18 and sets post-change coupling
to 0.26, 0.34, or 0.42. Across six models, $\Delta B$ rises from .00438 to .02917 whereas $\Delta D$
remains at or below .00119. Oracle distance accounts for 0.68--3.93\% of the decomposed
expected-risk increase, and all 18 simultaneous one-sided 95\% upper bounds are below 6.52\%.
In these linear-Gaussian VARs, stronger coupling increases the spectral radius and propagates innovation
uncertainty more strongly, providing the designed environmental channel through which $\Delta B$ rises.
The attribution therefore characterizes the evaluated controlled systems rather than arbitrary nonlinear
relation shifts.

A targeted replication changes both dimensionality and graph structure. In an independently generated
eight-channel two-community VAR, DLinear, PatchTST, and Chronos-2 yield fractions of 0.84--2.96\%.
These models were prespecified to represent linear target-trained, deep target-trained, and frozen
pretrained forecasting. All upper bounds are below 3.92\%. Appendix
Table~\ref{tab:stage4-system-comparison} characterizes the two
systems through their graph structures, spectral radii, and environmental-risk changes. Together, these
results assign most expected-risk degradation to greater environmental uncertainty, while
oracle distance contributes a comparatively small share.

Matching environmental-risk increase at approximately $\Delta B=0.01235$ provides a separate topology
audit. Across Ring, Block, and Hub relations, six models, and two times since shift, all 36 point estimates
assign a majority of degradation to environmental risk; 35 of 36 simultaneous upper bounds are below
10\%, and all are below 12\%. The larger fractions after 192 post-shift observations indicate
heterogeneous passive contextual response, not parameter adaptation. Complete values and the full
topology visualization appear in Appendix Table~\ref{tab:stage5-topology} and
Figure~\ref{fig:topology-audit}. In addition, Table~\ref{tab:quantitative-summary} consolidates calibration,
coupling attribution, and confirmation.

\subsection{Held-out stability of discovery profiles}

\begin{figure}[!htbp]
\centering
\includegraphics[width=\textwidth]{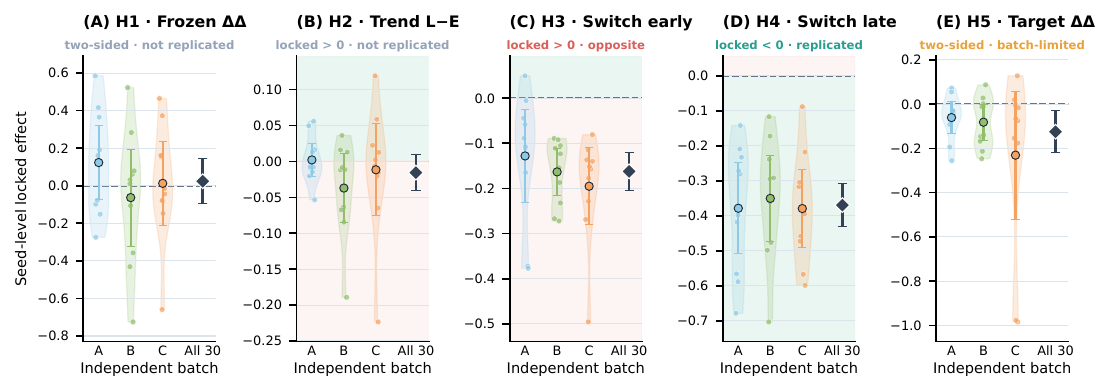}
\vspace{-0.8em}
\caption{\textbf{Held-out batches separate stable from visually compelling discovery patterns.}
Violin shapes and small points show seed-level distributions for the prespecified contrasts H1 through H5
in independent batches A, B, and C; larger circles and whiskers show batch means and two-sided 95\% Student-$t$
intervals, and dark diamonds show the All-30 precision analysis. Green and red regions mark the locked and
opposite directions for one-sided tests. The vertical axis is the seed-level value of each locked scalar
contrast; zero denotes no contrast, and panel scales differ. Only H4 is supported in every batch.}
\label{fig:confirmation-forest}
\end{figure}

\begin{figure}[H]
\centering
\includegraphics[width=0.92\textwidth]{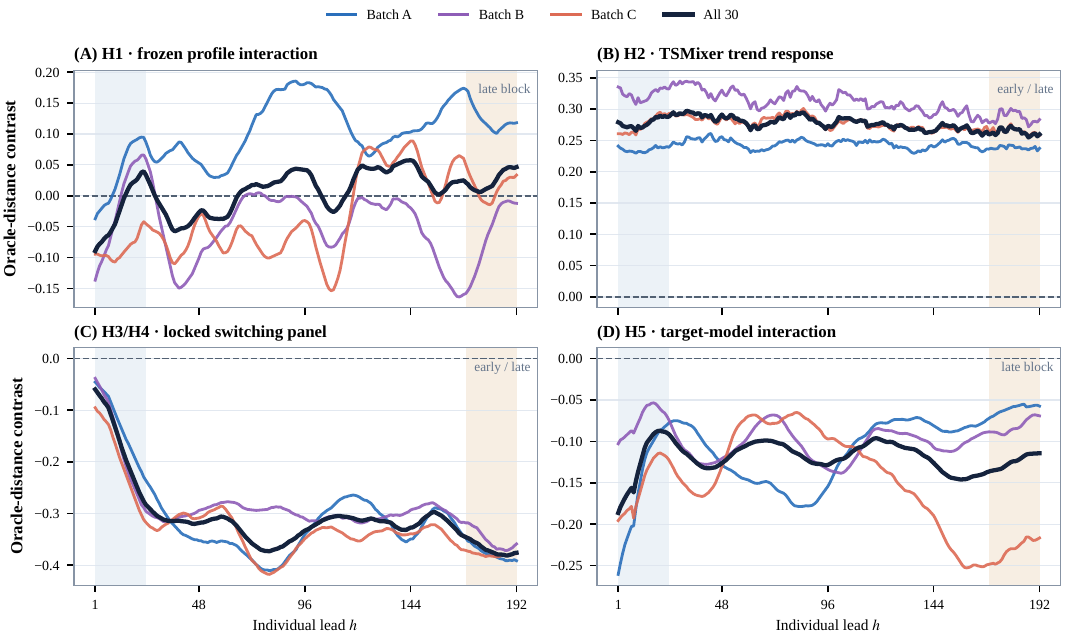}
\vspace{-0.7em}
\caption{\textbf{Held-out lead profiles reveal batch heterogeneity.} Rolling means compare batches A--C
with pooled seeds; shading marks locked blocks. Decisions follow Table~\ref{tab:confirmation}.}
\label{fig:heldout-batch-profiles}
\end{figure}

Figures~\ref{fig:confirmation-forest} and~\ref{fig:heldout-batch-profiles} separate locked scalar decisions
from their lead-resolved batch behavior. H4 is negative in all three batches, whereas H2 does not replicate
and H3 points opposite to its prespecified direction throughout. H1 also fails to replicate, and H5 is
direction-consistent but batch-limited. Complete definitions, multiplicity correction, and sign-flip
sensitivity appear in Appendix Table~\ref{tab:confirmation}; Table~\ref{tab:quantitative-summary}
consolidates the main numerical evidence.

Appendix Table~\ref{tab:switching-amplitude} shows early-block amplitude contraction from .550 to .261,
whereas late-block Base and Shift amplitudes are both .250. Thus differential late-block target contraction
does not explain the confirmed distance reduction, which may instead reflect how the evaluated procedures
approach the common long-run predictive center under faster switching.

\paragraph{Architecture-linked failure hypotheses.}
Model-resolved effects suggest trend extrapolation, stale persistence, shock propagation, and passive
context response as failure routes. AR avoids the TSMixer trend offset, whereas naive rules carry regimes
or shocks forward. Appendix~\ref{app:architecture-mechanisms} gives model-wise evidence and alternatives;
these remain structural hypotheses rather than causal ablations.

\paragraph{Interpretive boundary.}
The confirmed multivariate result attributes an expected-risk increase, not absolute quality. A small
$\Delta D$ can coexist with substantial error, and a negative contrast need not imply adaptation.
$\Delta B$ dominance limits model-only repair, whereas $\Delta D$ points to information or model
specification; intervention is required to establish repairability.

\FloatBarrier

\vspace{-9pt}

\section{Related work}

Monash, TFB, and GIFT-Eval broaden evaluation over observed datasets and models
\citep{godahewa2021monash,qiu2024tfb,aksu2024gifteval}; M4, M5, ProbTS, and fev-bench emphasize competition-scale,
horizon-aware, probabilistic, or aggregated comparisons
\citep{makridakis2020m4,makridakis2022m5,zhang2024probts,shchur2025fev}. SynTSBench and FinStressTS instead
isolate temporal or financial mechanisms~\citep{tan2025syntsbench,sun2026finstressts}. These settings broaden
coverage or isolate perturbations, but without an origin-conditioned target a loss increase still conflates
environmental difficulty with oracle distance; our paired oracle protocol supplies that attribution.

DomainBed standardizes domain generalization, while WILDS and Wild-Time curate natural cross-domain and
temporal shifts~\citep{gulrajani2021domainbed,koh2021wilds,yao2022wildtime}. They measure transfer; our
seed-matched counterfactual instead attributes expected squared-risk changes at individual leads. Work on
benchmark repositories, documentation, variance, and leakage cautions against treating leaderboards as
measurement guarantees~\citep{longjohn2024repositories,bhardwaj2024curation,bouthillier2021variance,paleka2026pitfalls}.
RevIN and methods for non-stationarity, dynamics, ensembling, or dependency shifts provide model-side
responses~\citep{kim2022revin,liu2022nonstationary,liu2023san,liu2023koopa,zhang2023onenet,ye2024fan,dai2024ddn,liu2025timebridge};
our fixed protocol compares them with statistical, pretrained, and target-trained methods.

For squared loss, the conditional mean is the Bayes point target; proper scores similarly link distributional
forecasts to explicit targets~\citep{gneiting2007strictly}. We operationalize this target through strict
\Base{}/\Shift{} pairing, origin conditioning, individual-lead profiles, full-pipeline controls, and held-out
confirmation. Appendix Table~\ref{tab:prior-scope} compares the resulting scope.

\begin{table}[H]
\caption{\textbf{Main results for calibration, coupling generalization, and held-out stability.}}
\label{tab:quantitative-summary}
\centering
\scriptsize
\setlength{\tabcolsep}{3.2pt}
\renewcommand{\arraystretch}{0.82}
\resizebox{\textwidth}{!}{%
\begin{tabular}{lllll}
\toprule
\rowcolor{bayesblue!32}
\multicolumn{5}{l}{\textbf{A. End-to-end controls with known attribution}} \\
\rowcolor{bayesblue!14}
Evidence & Scope & Reference quantity & Observed model term & Audit decision \\
\midrule
Identical-data null & 400 profiles & $\Delta B=0$ & $\Delta D=0$ & all 400 exactly zero \\
Innovation scale $.2\!\to\!.4$ & 100 seeds & $\Delta B=.120000$ & $f=.20\%$ & $U_{.95}=.22\%<10\%$ \\
Delete latest AR input & 100 seeds & $\Delta B=0$ & $\Delta D=.000315$; $f=100\%$ & 95\% CI $[.000262,.000372]$ \\
\midrule
\rowcolor{improvegreen!22}
\multicolumn{5}{l}{\textbf{B. Coupling attribution across severity and structure}} \\
\rowcolor{improvegreen!10}
Evidence & Scope & Environmental $\Delta B$ & Added-distance fraction $f$ & Upper-bound audit \\
\midrule
Ring $.18\!\to\!.26$ & 6 models, 20 seeds & $.00438$ & $.68$--$3.17\%$ & maximum $U_{.95}=4.30\%$ \\
Ring $.18\!\to\!.34$ & 6 models, 20 seeds & $.01235$ & $.71$--$3.39\%$ & maximum $U_{.95}=4.45\%$ \\
Ring $.18\!\to\!.42$ & 6 models, 20 seeds & $.02917$ & $.85$--$3.93\%$ & maximum $U_{.95}=6.52\%$ \\
8-channel communities & 3 models, 20 seeds & $.00311$ & $.84$--$2.96\%$ & maximum $U_{.95}=3.92\%$ \\
Matched Ring/Block/Hub & 6 models, 20 seeds each & $\approx.01235$ & $.23$--$9.05\%$ & 35/36 $U_{.95}<10\%$; all $<12\%$ \\
\midrule
\rowcolor{shiftorange!18}
\multicolumn{5}{l}{\textbf{C. Held-out stability on three independent ten-seed batches}} \\
\rowcolor{shiftorange!8}
Locked contrast & Scope & Direction & All-30 effect [95\% CI] & Batch-level decision \\
\midrule
H1 frozen-model interaction & A/B/C & two-sided & $.024\ [-.094,.143]$ & not replicated \\
H2 trend accumulation & A/B/C & $>0$ & $-.016\ [-.040,.009]$ & not replicated \\
H3 early switching & A/B/C & $>0$ & $-.162\ [-.205,-.120]$ & opposite in every batch \\
H4 late switching & A/B/C & $<0$ & $-.369\ [-.431,-.308]$ & supported in every batch \\
H5 target-model interaction & A/B/C & two-sided & $-.125\ [-.219,-.030]$ & direction-consistent; batch-limited \\
\bottomrule
\end{tabular}
}
\vspace{2pt}

\raggedright\scriptsize
$U_{.95}$ denotes the simultaneous one-sided 95\% upper confidence bound; fractions require positive total degradation.
\end{table}
\vspace{-10pt}

\section{Limitations and broader use}

Exact attribution requires known oracle moments and squared loss and does not establish observational
causality. Evidence covers ten-seed discovery and linear-Gaussian VARs, with only three methods in the
eight-channel system; one of 36 topology cells exceeds the 10\% simultaneous upper bound. Retraining,
fixed-weight OOD response, and post-shift adaptation are distinct estimands (Appendix~\ref{app:limitations}).
The 10\% margin is a prespecified diagnostic threshold rather than a universal constant, and the fraction
requires positive total degradation. Extension to nonlinear generators, other losses, and less controlled
shifts remains future work.

\section{Conclusion}

Average forecasting error shows whether performance changes, but not why. Our paired benchmark makes
that distinction operational by attributing expected squared-risk changes at every lead to environmental
risk and forecast--oracle distance, while reserving full-history oracle information for evaluation.
End-to-end null, environmental-only, and information-gap controls recover their known attribution. Across
the evaluated coupling severities and linear-Gaussian VAR structures, degradation is consistently dominated
by environmental risk rather than greater distance from the conditional mean. Held-out confirmation
provides a second safeguard by rejecting visually compelling discovery profiles that do not reproduce on
independent DGP realizations. Together, these results position the benchmark as both an attribution
instrument and a stability audit. It complements rather than replaces realistic OOD evaluation, which
remains necessary to establish practical behavior under naturally occurring shifts.

\newpage
\subsection*{AI use statement}

LLMs were used solely for language polishing and limited assistance with experimental code. They were not
used for research ideation, experimental design, data analysis, or interpretation of the results. All
AI-assisted outputs were reviewed and verified by the authors, who take full responsibility for the
manuscript and its findings.

\bibliography{references}
\bibliographystyle{iclr2027_conference}

\clearpage
\appendix
\setlength{\intextsep}{7pt plus 1pt minus 1pt}

\section{Prespecified held-out analyses and complete confirmatory results}
\label{app:lock}

The contrasts, directions, lead blocks, model panels, tests, and multiplicity families were specified in
timestamped local manifests before held-out execution. We refer to this evidence as a local prospective
lock, reserving ``preregistration'' for public third-party records. Each SHA-256 value in
Table~\ref{tab:lock-fingerprints} fingerprints the exact contents of the original prespecified manifest or protocol note:
the digest exposes subsequent edits and the timestamp records when the local file was frozen. The
Stage-2 scientific choices were frozen at 2026-09-10 20:41:40 Asia/Shanghai. The later v3 execution
manifest records only a uniform TimesFM OOM fallback and explicitly retains the earlier scientific-config
hashes. The Stage-2.5/3 extension also fixed the All-30 precision analysis before Stage-2 analysis was
generated, while the three batches remain separately visible. When an original manifest contains a
machine-specific path or host name, the anonymous artifact supplies an explicitly labeled redacted copy
with a separate release digest; these redactions do not alter the scientific protocol.

\begin{table}[H]
\caption{\textbf{Prospective protocol locks and file hashes.} SHA-256 fingerprints identify the exact bytes of each
original prespecified record. Readable prefixes are shown here; the anonymous artifact records the complete original and release-copy digests for
recomputation. Times use Asia/Hong Kong (UTC$+8$).}
\label{tab:lock-fingerprints}
\centering
\scriptsize
\setlength{\tabcolsep}{3pt}
\begin{tabular}{p{0.18\textwidth}p{0.13\textwidth}p{0.20\textwidth}p{0.42\textwidth}}
\toprule
Record & Frozen & SHA-256 prefix & Purpose \\
\midrule
Stage-2 execution v3 & Sep 10, 20:58 & \texttt{7951e4b02fa39a86} & Documents uniform OOM fallback; science unchanged \\
Stage-2.5/3 plan & Sep 10, 23:05 & \texttt{27c3500e69930ebb} & Locks new seeds, tasks, and All-30 analysis \\
Stage-3 clarification & Sep 10, 23:50 & \texttt{fb56eb02cf0f58bd} & Fixes estimands before model results \\
Stage-4 manifest & Sep 11, 13:32 & \texttt{50fadb5ec398f18f} & Locks controls, severity, and second system \\
Topology manifest & Sep 11, 15:27 & \texttt{bb0346f88d410ad5} & Locks matched Ring/Block/Hub audit \\
\bottomrule
\end{tabular}
\end{table}

Discovery uses seeds 1101--1110; held-out batches A, B, and C use 2101--2110, 2201--2210, and
2301--2310.  Stage-3A and Stage-3B use 3201--3220 and 3301--3320.  Stage-4 environmental, information-gap,
severity, and second-DGP controls use 4201--4300, 4301--4400, 4401--4420, and 4501--4520, respectively.
The topology audit uses independent Block seeds 4601--4620 and Hub seeds 4701--4720; the Ring panel is
the previously locked Stage-4 reference.
Complete seed-to-initialization mappings and file hashes are retained in the released manifests. H5 was
locked as a separate one-test family because it was the sole target-trained-protocol interaction introduced
at Stage 2. Its one-test family $p$-value and batch-level estimates are both reported.

\begin{table}[t]
\caption{Held-out results from three independent ten-seed batches. Two-sided 95\% Student-$t$ intervals quantify effect size. $p_{t,H}$ is the direction-matched Student-$t$ $p$-value after Holm correction; $p_{\rm SF,H}$ is the corresponding seed-level sign-flip sensitivity result. H1--H4 form one family within each batch and pooled analysis; H5 is its separately locked one-test family.}
\label{tab:confirmation}
\centering
\footnotesize
\setlength{\tabcolsep}{2.8pt}
\resizebox{\textwidth}{!}{%
\begin{tabular}{llccc}
\toprule
Hyp. & Alternative & A: effect [95\% CI]; $p_{t,H}$ & B: effect [95\% CI]; $p_{t,H}$ & C: effect [95\% CI]; $p_{t,H}$ \\
\midrule
H1 & two-sided & \shortstack{\(0.123\) [\(-0.074\), \(0.321\)]\\\(0.574\)} & \shortstack{\(-0.063\) [\(-0.322\), \(0.195\)]\\\(1.000\)} & \shortstack{\(0.013\) [\(-0.210\), \(0.235\)]\\\(1.000\)} \\
H2 & $>0$ & \shortstack{\(0.002\) [\(-0.021\), \(0.026\)]\\\(0.844\)} & \shortstack{\(-0.037\) [\(-0.085\), \(0.010\)]\\\(1.000\)} & \shortstack{\(-0.012\) [\(-0.075\), \(0.052\)]\\\(1.000\)} \\
H3 & $>0$ & \shortstack{\(-0.128\) [\(-0.230\), \(-0.026\)]\\\(0.990\)} & \shortstack{\(-0.163\) [\(-0.215\), \(-0.112\)]\\\(1.000\)} & \shortstack{\(-0.195\) [\(-0.280\), \(-0.110\)]\\\(1.000\)} \\
H4 & $<0$ & \shortstack{\(-0.379\) [\(-0.509\), \(-0.248\)]\\\(2.05\times 10^{-4}\)} & \shortstack{\(-0.350\) [\(-0.474\), \(-0.227\)]\\\(2.44\times 10^{-4}\)} & \shortstack{\(-0.379\) [\(-0.491\), \(-0.268\)]\\\(6.08\times 10^{-5}\)} \\
H5 & two-sided & \shortstack{\(-0.061\) [\(-0.133\), \(0.011\)]\\\(0.087\)} & \shortstack{\(-0.082\) [\(-0.163\), \(-2.13\times 10^{-4}\)]\\\(0.050\)} & \shortstack{\(-0.231\) [\(-0.520\), \(0.058\)]\\\(0.105\)} \\
\midrule
\multicolumn{5}{l}{\emph{Sign-flip sensitivity: Holm-adjusted $p_{\rm SF,H}$ (H5 unadjusted as a one-test family)}} \\
\midrule
Hyp. & Alternative & A & B & C \\
\midrule
H1 & two-sided & \(0.562\) & \(1.000\) & \(1.000\) \\
H2 & $>0$ & \(0.865\) & \(1.000\) & \(1.000\) \\
H3 & $>0$ & \(0.997\) & \(1.000\) & \(1.000\) \\
H4 & $<0$ & \(0.004\) & \(0.004\) & \(0.004\) \\
H5 & two-sided & \(0.084\) & \(0.051\) & \(0.057\) \\
\bottomrule
\end{tabular}
}
\vspace{2pt}

\begin{tabular}{lcl}
\toprule
Hyp. & All 30: effect [95\% CI]; $p_{t,H}$ / $p_{\rm SF,H}$ & Outcome \\
\midrule
H1 & \(0.024\) [\(-0.094\), \(0.143\)]; \(1.000\) / \(1.000\) & not replicated \\
H2 & \(-0.016\) [\(-0.040\), \(0.009\)]; \(1.000\) / \(1.000\) & not replicated \\
H3 & \(-0.162\) [\(-0.205\), \(-0.120\)]; \(1.000\) / \(1.000\) & opposite direction in A/B/C \\
H4 & \(-0.369\) [\(-0.431\), \(-0.308\)]; \(1.09\times 10^{-12}\) / \(4.00\times 10^{-6}\) & supported in A/B/C \\
H5 & \(-0.125\) [\(-0.219\), \(-0.030\)]; \(0.011\) / \(6.94\times 10^{-4}\) & direction-consistent; batch-limited \\
\bottomrule
\end{tabular}
\par\vspace{2pt}\raggedright\footnotesize For H3, the two-sided confidence intervals are entirely negative while the directional $p$-values are near one because the prospectively locked alternative was positive. All 30 was locked before held-out execution as a precision analysis and does not replace the three batch-specific results.
\end{table}

Table~\ref{tab:confirmation} preserves each ten-seed batch instead of presenting only the more precise
All-30 estimate. It shows that H4 is the only directional claim supported in A, B, and C; H1 and H2 are
not replicated, H3 consistently points opposite to its locked direction, and H5 is pooled-directional but
batch-limited. Student-$t$ and exact sign-flip sensitivity results agree on these decisions.

With $\bar A_m^k(a{:}b)=\operatorname{mean}_{h=a:b}A_m(k,h)$, the complete scalar hypotheses are
\begin{align}
{\rm H1}:~&
[\bar A_{\rm Chronos}^{\rm trend}-\bar A_{\rm Chronos}^{\rm switch}]
-[\bar A_{\rm TimesFM}^{\rm trend}-\bar A_{\rm TimesFM}^{\rm switch}]\ne0,\\
{\rm H2}:~&
\bar A_{\rm TSMixer}^{\rm trend}(169{:}192)
-\bar A_{\rm TSMixer}^{\rm trend}(1{:}24)>0,\\
{\rm H3}:~&\bar A_{\rm panel}^{\rm switch}(1{:}24)>0,\qquad
{\rm H4}:~\bar A_{\rm panel}^{\rm switch}(169{:}192)<0,\\
{\rm H5}:~&
[\bar A_{\rm SegRNN}^{\rm trend}-\bar A_{\rm SegRNN}^{\rm switch}]
-[\bar A_{\rm NST}^{\rm trend}-\bar A_{\rm NST}^{\rm switch}]\ne0,
\end{align}
where unmarked H1/H5 bars use leads 169--192 and the fixed H3/H4 panel is AR, Chronos-2, TimesFM,
and TSMixer.  The primary tests and correction scopes are stated in Section~\ref{sec:protocol}.

\begin{figure}[H]
\centering
\includegraphics[width=\textwidth]{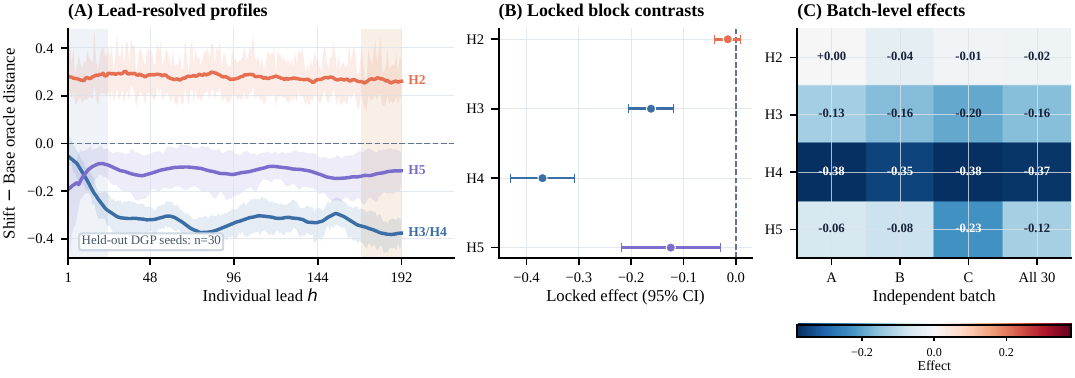}
\caption{\textbf{Lead-resolved and block-level held-out evidence for H2--H5.}
(A) overlays the three lead profiles on a common scale. (B) reports the four locked block contrasts with
95\% confidence intervals. (C) shows their batch-level means; confirmatory decisions still follow the
prespecified batch-level rules.}
\label{fig:horizon-profiles}
\end{figure}

Figure~\ref{fig:horizon-profiles} places the three commensurate lead profiles on one axis, then links them
to the locked block contrasts and independent-batch effects. Its role is diagnostic: the curve, interval,
and heatmap views explain where each scalar result originates, while Table~\ref{tab:confirmation} remains
the confirmatory decision record.

\IfFileExists{generated/stage3a/stage3a_summary.tex}{\begin{table}[H]
\caption{Attribution for a multivariate coupling-strength shift. $f_m=\Delta D_m/(\Delta B+\Delta D_m)$ uses analytic expected risk, rather than realized raw MSE, in its denominator.  Intervals and $U_{.95}$ are percentiles from 50,000 paired DGP-seed bootstrap draws; $U_{.95}$ is the one-sided 95\% upper bound.}
\label{tab:stage3a}
\centering
\small
\setlength{\tabcolsep}{3.2pt}
\begin{tabular}{lrrrrrr}
\toprule
Model & $\Delta$ MSE$_{\rm real}$ & $\Delta R_{\rm oracle}$ & $\Delta B$ & $\Delta D$ & $f_m$ [95\% CI] & $U_{.95}$ \\
\midrule
Chronos-2 & 0.01242 & 0.01263 & 0.01235 & 0.00028 & 0.022 [0.013, 0.033] & 0.031 \\
TTM & 0.01258 & 0.01264 & 0.01235 & 0.00029 & 0.023 [0.013, 0.035] & 0.033 \\
DLinear & 0.01250 & 0.01263 & 0.01235 & 0.00028 & 0.022 [0.017, 0.028] & 0.027 \\
PatchTST & 0.01275 & 0.01289 & 0.01235 & 0.00054 & 0.042 [0.034, 0.050] & 0.049 \\
TSMixer & 0.01227 & 0.01251 & 0.01235 & 0.00016 & 0.013 [0.008, 0.018] & 0.017 \\
iTransformer & 0.01223 & 0.01259 & 0.01235 & 0.00024 & 0.019 [0.013, 0.027] & 0.025 \\
\bottomrule
\end{tabular}
\end{table}
}{}
\enlargethispage{2\baselineskip}
\IfFileExists{generated/stage3a/stage3a_origin_summary.tex}{\begin{table}[H]
\caption{Attribution by forecast origin for the multivariate coupling-strength shift. Origin 3,712 uses an all-pre-change context; origin 3,904 includes 192 post-change observations without a parameter update. Entries are $f_m$ (one-sided 95\% upper bound).}
\label{tab:stage3a-origin}
\centering
\footnotesize
\begin{tabular}{lrr}
\toprule
Model & Origin 3,712 & Origin 3,904 \\
\midrule
Chronos-2 & 0.006 (0.009) & 0.038 (0.055) \\
TTM & 0.005 (0.008) & 0.040 (0.060) \\
DLinear & 0.009 (0.013) & 0.035 (0.043) \\
PatchTST & 0.005 (0.008) & 0.075 (0.088) \\
TSMixer & 0.010 (0.016) & 0.015 (0.024) \\
iTransformer & 0.009 (0.014) & 0.029 (0.042) \\
\bottomrule
\end{tabular}
\end{table}
}{}
\IfFileExists{generated/stage3b/stage3b_summary.tex}{\begin{table}[H]
\caption{Standardized Student-$t_5$ Shift versus Gaussian Base at $H=192$.}
\label{tab:stage3b}
\centering
\small
\begin{tabular}{lrrr}
\toprule
Model & $\Delta R_{oracle}$ & 95\% CI & $p_H$ \\ 
\midrule
Chronos-2 & 0.0020 & [-0.0000, 0.0041] & 0.324 \\
TimesFM & 0.0032 & [-0.0004, 0.0067] & 0.378 \\
AR & -0.0002 & [-0.0004, 0.0001] & 0.638 \\
NST & 0.0007 & [-0.0003, 0.0017] & 0.592 \\
SegRNN & 0.0002 & [-0.0001, 0.0004] & 0.638 \\
TSMixer & -0.0161 & [-0.1040, 0.0719] & 0.707 \\
\bottomrule
\end{tabular}
\end{table}
}{}
Tables~\ref{tab:stage3a} and~\ref{tab:stage3a-origin} give the absolute risk components and separate the
immediate post-change origin from the later passive-context origin. The environmental term dominates in
both views, although the later origin permits a larger forecast--oracle-distance share for several models.
Table~\ref{tab:stage3b} is a distinct finite-fourth-moment heavy-tail check: all six intervals span zero.
Because this check was designed for effect detection rather than equivalence testing, it leaves the size
of any small residual effect unresolved.
\FloatBarrier
\IfFileExists{generated/stage4/severity_table.tex}{\begin{table}[H]
\caption{\textbf{Coupling attribution at three shift severities.} Cells are added oracle-distance fraction in percent, with the locked family-wise one-sided 95\% upper bound in parentheses. Bounds use 50,000 paired DGP-seed bootstrap draws and Bonferroni correction over the 18 cells. The common Base coupling is 0.18; 20 new DGP seeds are used.}
\label{tab:stage4-severity}
\centering
\footnotesize
\setlength{\tabcolsep}{5pt}
\begin{tabular}{lrrr}
\toprule
Model & 0.26 & 0.34 & 0.42 \\
\midrule
Chronos-2 & 1.05 (1.65) & 1.25 (2.43) & 2.33 (6.52) \\
TTM & 1.46 (2.37) & 1.64 (2.60) & 2.01 (3.41) \\
DLinear & 1.68 (2.11) & 1.74 (2.18) & 1.94 (2.54) \\
PatchTST & 3.17 (4.30) & 3.39 (4.45) & 3.93 (5.15) \\
TSMixer & 0.68 (1.03) & 0.71 (1.04) & 0.85 (1.41) \\
iTransformer & 1.14 (1.77) & 1.20 (1.90) & 1.46 (2.37) \\
\bottomrule
\end{tabular}
\end{table}
}{}
\IfFileExists{generated/stage4/second_dgp_table.tex}{\begin{table}[H]
\caption{\textbf{Attribution in the eight-channel two-community VAR.} Cross-community coupling changes from 0.02 to 0.16 over 20 new DGP seeds. $U_{.95}$ uses 50,000 paired DGP-seed bootstrap draws and Bonferroni correction over the three models.}
\label{tab:stage4-second-dgp}
\centering
\footnotesize
\setlength{\tabcolsep}{4pt}
\begin{tabular}{lrrrr}
\toprule
Model & $\Delta B$ & $\Delta D$ & $f$ (\%) & $U_{.95}$ (\%) \\
\midrule
DLinear & 0.003110 & 0.000041 & 1.29 & 1.61 \\
PatchTST & 0.003110 & 0.000095 & 2.96 & 3.92 \\
Chronos-2 & 0.003110 & 0.000026 & 0.84 & 2.19 \\
\bottomrule
\end{tabular}
\end{table}
}{}
\IfFileExists{generated/stage4/system_comparison_table.tex}{\begin{table}[H]
\caption{\textbf{Configurations of the two multivariate systems.} Coupling coefficients are structure-specific controls rather than a common severity scale.  The eight-channel panel was fixed before execution to represent a linear target-trained model (DLinear), a deep target-trained model (PatchTST), and a frozen TSFM (Chronos-2).}
\label{tab:stage4-system-comparison}
\centering
\footnotesize
\setlength{\tabcolsep}{4pt}
\resizebox{\textwidth}{!}{%
\begin{tabular}{lcllllrr}
\toprule
System & Channels & Graph & Coupling Base$\to$Shift & Spectral radius Base$\to$Shift & $\Delta B$ & Models & Seeds \\
\midrule
Ring VAR & 4 & directed cycle & $.18\to\{.26,.34,.42\}$ & $.63\to\{.71,.79,.87\}$ & $.00438,.01235,.02917$ & 6 & 20 \\
Two-community VAR & 8 & two dense blocks & $.02\to.16$ & $.54\to.68$ & $.00311$ & 3 & 20 \\
\bottomrule
\end{tabular}
}
\end{table}
}{}
\IfFileExists{generated/final_advice/stage5_topology_table.tex}{\begin{table}[H]
\caption{\textbf{Attribution across coupling topologies with matched environmental difficulty.} Cells report the added-distance fraction in percent and the simultaneous one-sided 95\% upper bound in parentheses. Ring is the previously locked reference; Block and Hub use independent 20-seed batches with environmental-risk changes matched to the Ring value. Bold marks Block--PatchTST at $\tau=192$, the only bound above the prespecified 10\% margin (11.97\%).}
\label{tab:stage5-topology}
\centering
\footnotesize
\setlength{\tabcolsep}{3.4pt}
\begin{tabular}{lrrrrrr}
\toprule
& \multicolumn{2}{c}{Ring} & \multicolumn{2}{c}{Block} & \multicolumn{2}{c}{Hub} \\
\cmidrule(lr){2-3}\cmidrule(lr){4-5}\cmidrule(lr){6-7}
Model & $\tau=0$ & $\tau=192$ & $\tau=0$ & $\tau=192$ & $\tau=0$ & $\tau=192$ \\
\midrule
Chronos-2 & 0.44 (0.72) & 2.04 (4.45) & 0.30 (0.43) & 2.92 (5.04) & 0.84 (1.42) & 0.69 (3.56) \\
TTM & 0.55 (0.81) & 2.71 (4.63) & 0.23 (0.46) & 5.78 (9.84) & 0.86 (1.51) & 3.59 (8.28) \\
DLinear & 0.61 (1.07) & 2.85 (3.72) & 0.44 (0.70) & 4.10 (5.61) & 0.97 (1.67) & 2.80 (3.61) \\
PatchTST & 0.61 (1.04) & 6.02 (8.04) & 0.55 (0.92) & \textbf{9.05 (11.97)} & 1.11 (2.02) & 6.11 (8.17) \\
TSMixer & 0.61 (0.95) & 0.80 (1.54) & 0.62 (0.92) & 1.79 (3.12) & 1.11 (2.20) & 1.17 (2.21) \\
iTransformer & 0.46 (0.67) & 1.93 (3.37) & 0.59 (0.91) & 3.31 (4.44) & 1.15 (2.40) & 2.82 (5.05) \\
\bottomrule
\end{tabular}
\end{table}
}{}
Table~\ref{tab:stage4-severity} shows the six-model severity curve, and
Table~\ref{tab:stage4-second-dgp} tests a prespecified three-model panel on the eight-channel system.
Table~\ref{tab:stage4-system-comparison} reports the structure-specific coefficients beside their induced
risk changes, supporting comparison on a common risk scale. Finally, Table~\ref{tab:stage5-topology}
matches environmental-risk change across
Ring, Block, and Hub relations: 35 of 36 simultaneous upper bounds remain below the 10\% diagnostic
margin; the remaining Block--PatchTST bound at $\tau=192$ is 11.97\%.

\begin{figure}[tbp]
\centering
\includegraphics[width=0.97\textwidth]{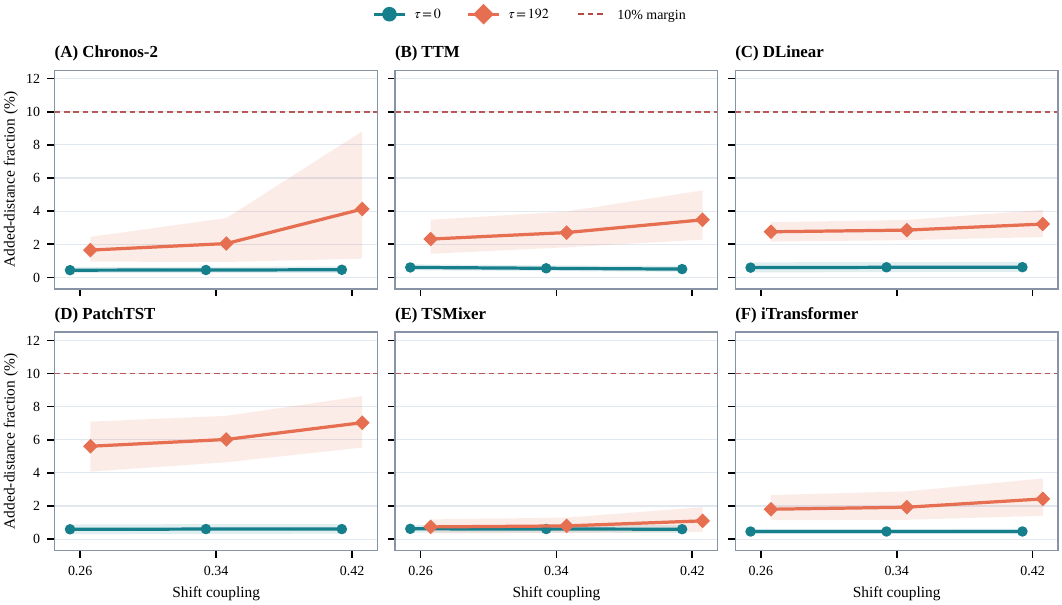}
\caption{\textbf{Origin-resolved coupling-severity responses for all six models.} Lines show the
added-distance fraction across the three Shift couplings; shaded bands are two-sided 95\% paired-seed
bootstrap intervals. The dashed line is the prespecified 10\% diagnostic margin, not a theoretical
constant.}
\label{fig:severity-by-model}
\end{figure}

Figure~\ref{fig:severity-by-model} resolves the severity experiment by model and time since shift. The
immediate-origin fractions remain tightly concentrated below 1\%, while the later origin reveals the
model-dependent contextual response that is hidden by origin averaging. The increase is most visible for
PatchTST and at the strongest coupling for Chronos-2; no parameter update occurs at either origin.

\FloatBarrier
\IfFileExists{generated/next_round/switching_oracle_amplitude.tex}{\begin{table}[H]
\caption{Oracle-target amplitude for the locked switching contrast. The analysis uses 30 held-out DGP seeds, with origins averaged within each seed. Entries are $\mathbb E[(\mu^F)^2]$ with two-sided 95\% Student-$t$ intervals.}
\label{tab:switching-amplitude}
\centering
\small
\begin{tabular}{lrrr}
\toprule
Lead block & Base & Shift & Shift$-$Base \\
\midrule
1--24 & 0.550 [0.410, 0.690] & 0.261 [0.248, 0.274] & -0.289 [-0.425, -0.153] \\
169--192 & 0.250 [0.250, 0.250] & 0.250 [0.250, 0.250] & -0.000 [-0.000, 0.000] \\
\bottomrule
\end{tabular}
\end{table}
}{}
Table~\ref{tab:switching-amplitude} tests a competing explanation for the negative late-switching
distance. The predictable target contracts strongly in the early block and is already at the same
long-run amplitude in Base and Shift late in the horizon. Therefore, differential late-block target
contraction is not a direct explanation of H4. The fixed model parameters rule out online parameter
adaptation; convergence toward the common long-run predictive center remains a plausible but unconfirmed
explanation.
\FloatBarrier

\section{Protocol and terminology details}

\begin{table*}[h]
\caption{Evaluation scope of representative forecasting benchmarks.  ``Partial'' denotes a
related capability with a different target; ``--'' means outside the work's primary scope, not a quality
judgment.  Entries describe the cited papers' documented primary evaluations.}
\label{tab:prior-scope}
\centering
\small
\setlength{\tabcolsep}{2.5pt}
\resizebox{\textwidth}{!}{%
\begin{tabular}{lccccccc}
\toprule
Benchmark & \shortstack{Real-data\\breadth} & \shortstack{Controlled\\mechanisms} & \shortstack{Strict\\pairing} & \shortstack{Origin\\oracle} & \shortstack{Risk\\split} & \shortstack{$k\!\times\!h$\\profile} & Adaptation \\
\midrule
TFB / GIFT-Eval~\citep{qiu2024tfb,aksu2024gifteval} & Yes & -- & -- & -- & -- & -- & -- \\
SynTSBench~\citep{tan2025syntsbench} & -- & Yes & -- & Partial & -- & Partial & -- \\
FinStressTS~\citep{sun2026finstressts} & -- & Yes & Partial & Partial & -- & Partial & -- \\
Ours & Partial & Yes & Yes & Yes & Yes & Yes & Yes \\
\bottomrule
\end{tabular}
}
\end{table*}

\begin{table*}[h]
\caption{Training and evaluation protocols.  The multivariate relation shift is separated from
the univariate environment-specific training protocol.}
\label{tab:protocols}
\centering
\small
\setlength{\tabcolsep}{3pt}
\begin{tabular}{>{\raggedright\arraybackslash}p{0.16\textwidth}
>{\raggedright\arraybackslash}p{0.27\textwidth}
>{\raggedright\arraybackslash}p{0.25\textwidth}
>{\raggedright\arraybackslash}p{0.24\textwidth}}
\toprule
Protocol & \Base{} & \Shift{} & What the contrast supports \\
\midrule
Statistical anchors & Refit from visible \Base{} context & Refit from visible \Shift{} context & Stress response of a fixed fitting rule \\
Univariate target-trained (15) & Train/validate/test on \Base{} (70/10/20) & Separately train/validate/test on \Shift{} (70/10/20) & Mechanism-controlled response of an environment-specific fitting procedure \\
Multivariate target-trained & Fit on common pre-change prefix & Identical pre-change data/seed; no post-change update & Immediate OOD at the change point; passive contextual response at the later origin \\
Frozen TSFM (5) & Fixed pretrained checkpoint & The identical checkpoint; only context changes & Fixed-model cross-environment stress \\
Matched adaptation (5) & Random/zero/adapt arms on \Base{} support/test & Separately matched arms on \Shift{} support/test & Whether environment-specific limited-data intervention repairs the diagnosed gap \\
\bottomrule
\end{tabular}
\end{table*}

\IfFileExists{generated/final_advice/all_model_aggregate.tex}{\begin{table*}[t]
\caption{\textbf{Aggregate results on the shared discovery grid.} The table contains 24 deployable forecasters plus the evaluator-only DGP oracle-mean reference. Lower is better within each metric, but raw MSE, oracle distance, and SMSE answer different questions. Values are descriptive averages over the common 38 case--horizon cells; DGP seed remains the independent unit for inference.}
\label{tab:all-model-aggregate}
\centering
\small
\setlength{\tabcolsep}{7pt}
\begin{tabular}{llrrr}
\toprule
Entry & Protocol & Raw MSE & Oracle distance & SMSE \\
\midrule
DGP oracle mean & Evaluator reference & 0.2558 & 0.0000 & 0.9948 \\
AR & Statistical & 0.2700 & 0.0159 & 1.0148 \\
ETS & Statistical & 0.3339 & 0.0912 & 1.1516 \\
Naive & Statistical & 0.4464 & 0.2185 & 1.8908 \\
Seasonal naive & Statistical & 0.5994 & 0.3509 & 2.4151 \\
SegRNN & Target-trained & 0.2672 & 0.0063 & 1.0213 \\
Non-stationary Tr. & Target-trained & 0.2877 & 0.0207 & 1.0537 \\
TimeKAN & Target-trained & 0.3036 & 0.0266 & 1.0702 \\
Autoformer & Target-trained & 0.2847 & 0.0335 & 1.1549 \\
Koopa & Target-trained & 0.3086 & 0.0451 & 1.1867 \\
TimeMixer & Target-trained & 0.3113 & 0.0488 & 1.1904 \\
TimesNet & Target-trained & 0.3265 & 0.0569 & 1.1529 \\
TexFilter & Target-trained & 0.3309 & 0.0709 & 1.1571 \\
DLinear & Target-trained & 0.3459 & 0.0765 & 1.2594 \\
iTransformer & Target-trained & 0.3383 & 0.0772 & 1.1756 \\
PaiFilter & Target-trained & 0.3427 & 0.0845 & 1.1717 \\
SAN-DLinear & Target-trained & 0.3580 & 0.0873 & 1.3311 \\
PatchTST & Target-trained & 0.3777 & 0.1088 & 1.2677 \\
FAN-DLinear & Target-trained & 0.3914 & 0.1272 & 1.3943 \\
TSMixer & Target-trained & 0.5662 & 0.3035 & 4.3769 \\
Chronos-2 & Frozen TSFM & 0.2917 & 0.0476 & 1.0426 \\
TimesFM & Frozen TSFM & 0.2963 & 0.0531 & 1.0900 \\
Moirai & Frozen TSFM & 0.3161 & 0.0723 & 1.1755 \\
TTM & Frozen TSFM & 0.3366 & 0.0827 & 1.8272 \\
Time-MoE & Frozen TSFM & 0.3356 & 0.0861 & 2.1265 \\
\bottomrule
\end{tabular}
\end{table*}
}{}

Table~\ref{tab:all-model-aggregate} supplies the complete aggregate audit behind any abbreviated model
discussion: all 24 deployable forecasters and the evaluator-only oracle reference appear under the same
38-cell discovery intersection. The three columns are intentionally not collapsed because they measure
operational error, distance from the predictive center, and scale-relative error, respectively.

\begin{table*}[h]
\caption{Terms used throughout the paper.}
\label{tab:terms}
\centering
\small
\resizebox{\textwidth}{!}{%
\begin{tabular}{>{\raggedright\arraybackslash}p{0.19\textwidth}
>{\raggedright\arraybackslash}p{0.27\textwidth}
>{\raggedright\arraybackslash}p{0.23\textwidth}
>{\raggedright\arraybackslash}p{0.23\textwidth}}
\toprule
Term & Definition & Interpretation & Interpretive scope \\
\midrule
Raw risk & $\mathbb E(Y-\hat Y)^2$ & Operational squared loss & Jointly reflects environmental risk and oracle distance \\
Environmental risk & $\mathbb E\operatorname{Var}(Y\mid\mathcal F_t^{\mathrm{eval}})$ & Evaluator-conditioned uncertainty & Uses declared pre-origin information only \\
Forecast--oracle distance / $\MSEmu$ & $\mathbb E(\hat Y-\mu^F)^2$ & Information restriction plus model approximation/estimation & Combined distance, not architecture error alone \\
SMSE & $\mathbb E[(Y-\hat Y)^2/\max(\sigma^F,\epsilon)^2]$ & Mean pointwise error relative to local predictive scale & Secondary diagnostic, not MASE or probabilistic calibration \\
Stress profile & $\{A_m(k,h)\}_{k,h}$ & Paired response over mechanism and individual lead & Aggregation requires deployment weights over $(k,h)$ \\
\bottomrule
\end{tabular}
}
\end{table*}

Tables~\ref{tab:prior-scope} and~\ref{tab:protocols} separate benchmark capability from training--test
relations, while Table~\ref{tab:terms} fixes the vocabulary used to interpret every result. In particular,
forecast--oracle distance is the combined contribution of information restriction and
approximation/estimation.

\FloatBarrier
\section{Complete discovery-profile visual audit}

\begin{figure}[H]
\centering
\includegraphics[width=\textwidth]{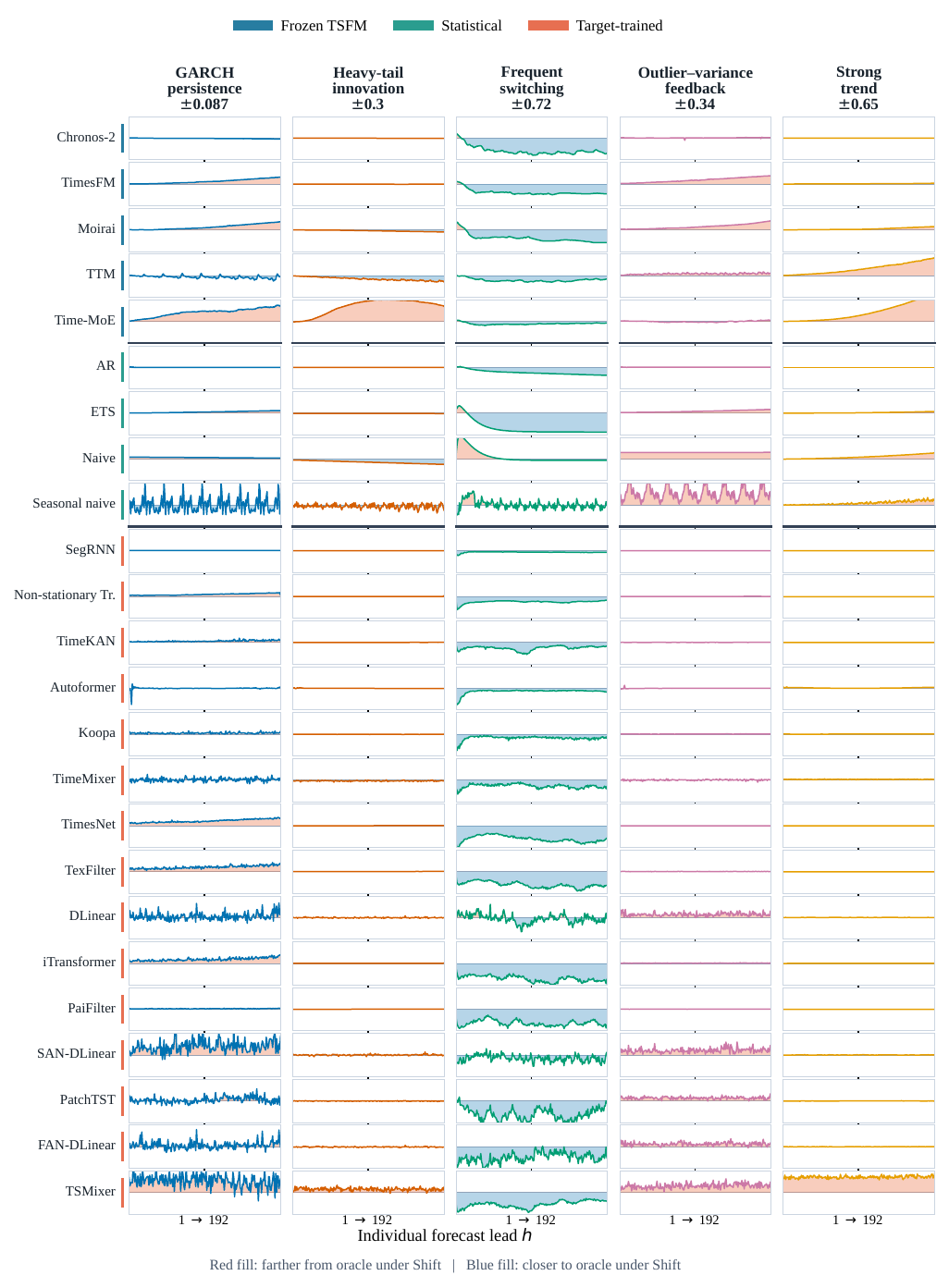}
\caption{\textbf{Complete lead-resolved forecast--oracle-distance atlas.} Rows are the 24 deployable
forecasters, columns are mechanism contrasts, and each microplot runs from lead 1 to 192; red/blue fill
denotes farther from/closer to the oracle under Shift. Column-specific symmetric limits are stated above
the panels, and protocol groups are separated by dark rules.}
\label{fig:appendix-sparkline-matrix}
\end{figure}

Figure~\ref{fig:appendix-sparkline-matrix} exposes the full individual-lead discovery atlas rather than a
selected set of visually attractive curves. Every row is one deployable forecaster and every available
cell is its Base-to-Shift change in forecast--oracle distance. A common vertical scale is used within each
mechanism column (the symmetric limit is printed in the header), so shapes, signs, and amplitudes are
comparable down a column. Each column has its own scale. The coverage audit confirms that all 24 methods
have all five $H=192$ individual-lead profiles and exactly the same 38 case--horizon cells in the aggregate
table.

\begin{figure}[H]
\centering
\includegraphics[width=0.97\textwidth]{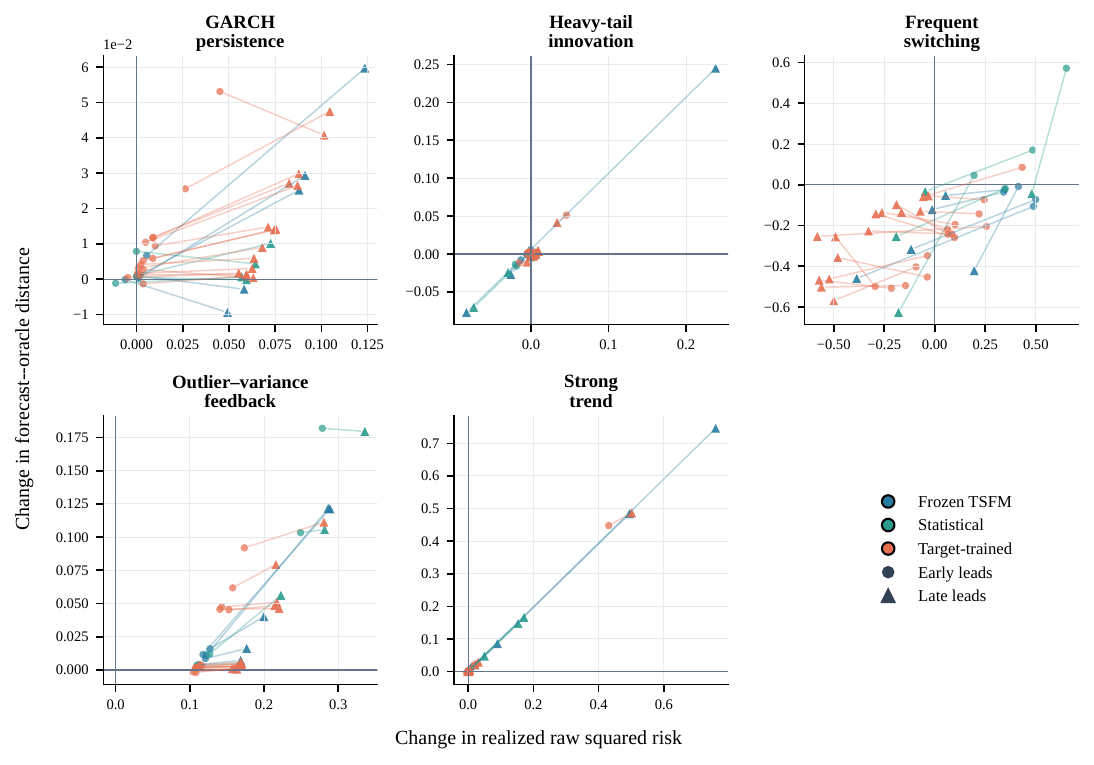}
\caption{\textbf{Early-to-late diagnostic phase portraits.} Circles average leads 1--24, triangles average
leads 169--192, and line segments connect the same forecaster; the horizontal and vertical axes are paired
changes in realized raw squared risk and forecast--oracle distance.}
\label{fig:appendix-phase-portraits}
\end{figure}

Figure~\ref{fig:appendix-phase-portraits} provides a complementary compression. It plots the change in
realized raw squared risk against the change in forecast--oracle distance for early and late lead blocks,
then connects the two summaries for each method. Points away from the diagonal reveal that raw-risk
movement cannot be read as forecast--oracle movement; trajectories that cross an axis show a horizon-dependent
diagnostic reversal. These discovery-stage portraits provide a compact visual screen whose selected
patterns are evaluated separately on held-out seeds.

\section{Architecture-linked interpretation of observed failure modes}
\label{app:architecture-mechanisms}

The controlled mechanisms reveal where a forecasting procedure fails, while model structure suggests how
that failure may arise. Table~\ref{tab:architecture-mechanisms} separates the measured pattern from the
structural interpretation. The first three rows use the ten-seed discovery grid, so their model-specific
intervals are descriptive cluster-bootstrap intervals without family-wise correction. The final row uses
the independent matched-difficulty topology audit. These comparisons identify plausible failure routes,
but models also differ in optimization, pretraining, and hyperparameters; consequently, the table does not
assign causality to an architectural component in isolation.

\begin{table}[t!]
\caption{\textbf{Observed failure patterns and architecture-linked hypotheses.} Numerical effects are changes in
forecast--oracle distance unless a fraction is stated. The evidence column reports observations, while the
interpretation column states the corresponding structural hypothesis.}
\label{tab:architecture-mechanisms}
\centering
\scriptsize
\setlength{\tabcolsep}{3.0pt}
\renewcommand{\arraystretch}{1.05}
\begin{tabular}{>{\raggedright\arraybackslash}p{0.12\textwidth}
>{\raggedright\arraybackslash}p{0.29\textwidth}
>{\raggedright\arraybackslash}p{0.29\textwidth}
>{\raggedright\arraybackslash}p{0.22\textwidth}}
\toprule
Stress & Model-resolved evidence & Structure-linked interpretation & Evidential status \\
\midrule
Strong trend & AR is numerically unchanged. TSMixer increases by $.305$ $[.281,.330]$, TTM by $.077$
$[.071,.085]$, and Time-MoE by $.086$ $[.073,.100]$; PatchTST changes by $-.003$
$[-.004,-.002]$. & AR explicitly fits a linear trend in each environment. Persistence and fixed-window
mixing do not enforce slope extrapolation, which can produce a level-dependent direct forecast. PatchTST
shows that lacking an explicit trend term is not sufficient for failure. & The TSMixer failure is clear in
discovery, but the locked H2 claim about stronger late-lead accumulation does not replicate. \\
Frequent switching & Descriptively, Naive increases by $.258$ $[.127,.376]$, whereas AR, ETS, TSMixer,
and Chronos-2 change by $-.071$, $-.172$, $-.421$, and $-.187$; ETS is not in the locked panel. The formal
H3/H4 panel is AR, Chronos-2, TimesFM, and TSMixer, whose late effect is negative in all three held-out
batches. & A last-value predictor persists the current regime and becomes stale after rapid switches.
Other forecasting procedures may approach the common long-run predictive center more closely under faster
switching. & The panel-level late reduction is confirmed. Its mechanism and individual-model differences
remain descriptive. \\
Outlier-variance feedback & Nineteen of 24 methods combine higher raw MSE with lower SMSE. Seasonal naive and
naive increase their oracle distance by $.182$ and $.104$, while AR changes by only $.003$. & Copying a
recent or seasonal observation can propagate an event-contaminated value. A fitted low-order dynamic can
dampen that transient. Much of the raw-loss increase still reflects predictive scale rather than model
distance. & The cross-protocol direction is broad, but the proposed propagation route has not been isolated
by an architectural ablation. \\
Relation shift & At $\tau=0$, every Base and Shift forecast is bitwise identical. At $\tau=192$, PatchTST
fractions are $6.02\%$, $9.05\%$, and $6.11\%$ for Ring, Block, and Hub, compared with $0.80\%$,
$1.79\%$, and $1.17\%$ for TSMixer. & Because weights are fixed, the later separation reflects how each
architecture converts 192 post-shift observations into a forecast. Patch tokenization and temporal mixing
are associated with different passive context responses, not different online adaptation. & The fractions
replicate on independent topology seeds; which internal operation causes the difference remains open. \\
\bottomrule
\end{tabular}
\end{table}

\paragraph{Trend extrapolation.}
The trend contrast most clearly separates an explicit parametric inductive bias from a learned direct
forecast. AR includes a constant and trend and is refit in each environment, so the stronger slope does not
create added oracle distance. TSMixer instead develops a large offset over the forecast window. This is
consistent with the fitted mixer using level-conditioned interpolation without a protected linear
extrapolation path. However, PatchTST and TimesNet remain near zero under the same training protocol.
Therefore, the evidence supports a failure of the evaluated TSMixer configuration rather than a universal
failure of nonlinear or patch-based models. The held-out result further narrows the statement because it
does not support the proposed increase from the early to the late lead block.

\paragraph{Regime persistence and long-run convergence.}
The switching results distinguish two routes to apparent failure. Naive carries the last observed regime
level forward, so a higher transition rate makes that state stale more quickly and increases oracle
distance. Several smoothing and learned forecasters move in the opposite direction. The oracle audit shows
early-block amplitude contraction, but the late Base and Shift amplitudes are both .250. Differential
late-block contraction therefore does not explain H4 directly. A plausible alternative is that these
procedures approach the common long-run predictive center differently under faster switching; identifying
that route requires an intervention. The negative distance effect is not evidence that a model detected the
switch or adapted its parameters.

\paragraph{Event propagation and scale.}
Outlier-variance feedback raises environmental scale after an outlier. The divergence between raw MSE and SMSE
across 19 methods follows directly from this scale change, so it should not be narrated as a widespread
architectural improvement. The larger distance changes for naive rules are nevertheless structurally
informative because direct copying preserves contaminated observations, whereas the fitted AR response is
strongly damped. This interpretation concerns transient propagation and remains distinct from the dominant
environmental-risk contribution.

\paragraph{Passive response to cross-channel change.}
The multivariate protocol provides the cleanest separation between parameter learning and context use.
Every model receives the same pre-change fitting data, and Base and Shift forecasts are identical at the
change point. At the later origin, the context contains 192 post-change observations but the parameters
remain fixed. The repeated PatchTST and TSMixer separation across matched Ring, Block, and Hub systems
therefore concerns passive context processing. It motivates future ablations of patch length, receptive
field, and channel mixing, while the present evidence establishes the response difference rather than the
specific internal cause.

\section{Proof of the conditional risk decomposition}
\label{app:decomp-proof}

Fix an environment, mechanism, forecast origin, and lead time, and abbreviate
$Y=Y^e_{t+h-1}$, $\hat Y=\hat Y^e_{m,t+h-1}$, and
$\mu=\mathbb E[Y\mid\mathcal F_t^{\mathrm{eval}}]$.  Assume finite second moments and that the trained
forecaster and its random seed are fixed, so $\hat Y$ is $\mathcal F_t^{\mathrm{eval}}$-measurable. If
prediction itself is randomized, the
same argument applies after additionally conditioning on that randomness.  Adding and subtracting $\mu$
gives
\begin{equation}
Y-\hat Y=(Y-\mu)+(\mu-\hat Y).
\end{equation}
Squaring and taking the conditional expectation yields
\begin{align}
\mathbb E[(Y-\hat Y)^2\mid\mathcal F_t^{\mathrm{eval}}]
&=\mathbb E[(Y-\mu)^2\mid\mathcal F_t^{\mathrm{eval}}]
 +2(\mu-\hat Y)\mathbb E[Y-\mu\mid\mathcal F_t^{\mathrm{eval}}]
 +(\mu-\hat Y)^2 \\
&=\operatorname{Var}(Y\mid\mathcal F_t^{\mathrm{eval}})+(\hat Y-\mu)^2,
\end{align}
because $\mathbb E[Y-\mu\mid\mathcal F_t^{\mathrm{eval}}]=0$.  Restoring the indices gives
Eq.~\ref{eq:decomp}.  The identity holds separately at every lead time; averaging over origins or lead
times preserves it. Its estimand is conditional expected squared risk; a single realized squared error is
one draw from that conditional risk.

For the nested information sets $\mathcal G_t^m\subseteq\mathcal F_t^{\mathrm{eval}}$, define
$\mu^G=\mathbb E[Y\mid\mathcal G_t^m]$.  The tower property gives
$\mathbb E[\mu^F\mid\mathcal G_t^m]=\mu^G$.  Expanding
$\mu^F-\hat Y=(\mu^F-\mu^G)+(\mu^G-\hat Y)$ and taking expectations makes the cross term zero because
$\mu^G-\hat Y$ is $\mathcal G_t^m$-measurable.  Hence
\begin{equation}
\mathbb E(\mu^F-\hat Y)^2
=\mathbb E(\mu^F-\mu^G)^2+\mathbb E(\mu^G-\hat Y)^2,
\end{equation}
which proves Eq.~\ref{eq:information-gap}. This is why $\MSEmu$ is called forecast--oracle distance rather than
pure model excess risk.

\section{Analytic means and Monte Carlo audit}
\label{app:mc-audit}

The first moment is analytic for every formal univariate DGP. For linear-mean, innovation, GARCH,
additive-outlier, and outlier-variance-feedback cases it is the deterministic center $a+b(t+h-1)$.
The supplemental stochastic-volatility case uses the same center and the recorded pre-origin latent scale
only for its scale oracle. For Markov switching, pre-origin observations give filtered state probabilities $\pi_t$ and
$\mu^F_{t,h}=\pi_tP^{h-1}\mathbf a$, where $\mathbf a$ contains the state means.  For symmetric persistent
level shocks, an observed pre-origin level
$L_{t-1}$ contributes $\rho^hL_{t-1}$ because future signed shocks have zero mean. These formulas replace the
Monte Carlo mean in all primary univariate $\MSEmu$ results.

For sampled futures, the evaluator estimates predictive scale by
\begin{equation}
\hat\sigma^{e,(i)}_{t,h}=\left[\frac{1}{S}\sum_{s=1}^{S}
\left(Y^{e,(i,s)}_{t+h-1}-\hat\mu^{e,(i)}_{t,h}\right)^2\right]^{1/2},
\label{eq:oracle-scale}
\end{equation}
the population standard deviation of the path ensemble (\texttt{ddof=0}).

If $\hat\mu=S^{-1}\sum_{s=1}^S Y^{(s)}$ is used instead, independence of the fixed forecast and evaluator
paths yields
\begin{equation}
\mathbb E_{\mathrm{MC}}(\hat Y-\hat\mu)^2
=(\hat Y-\mu^F)^2+v^F/S.
\end{equation}
Writing $\hat\sigma_0^2=(\hat\sigma^{e,(i)}_{t,h})^2$, the stored $\texttt{ddof}=0$ scale gives
$\hat\sigma_0^2/(S-1)$ as an unbiased estimator of $v^F/S$. We retain
this corrected estimator as an audit and use independent $S=2048,4096,8192$ ensembles to assess the scale
used by SMSE.

Across the formal $H=192$ windows, the implied $v^F/S$ bias ranges from $2.0\times10^{-5}$ to
$5.07\times10^{-4}$, comparable to the smallest empirical attribution effects and therefore not safely
ignorable.  The observed Monte Carlo mean RMSE tracks its estimated standard error.  Relative to 8,192
paths, the 2,048-path scale has 0.38--5.82\% relative MAE across the ten selected environments; the
4,096-path maximum is 3.58\%.  These findings motivate analytic means while retaining the path-count
sensitivity for scale-based metrics.

\paragraph{Evaluator controls already completed.}
Three control types are complete. A 100-seed identical-data null runs Naive, AR, Chronos-2, and TimesFM through
full prediction and evaluation (including real frozen-model inference). Across 400 model--seed profiles,
the maximum absolute raw-MSE, oracle-distance, or SMSE effect is zero, with 0/120 model-level Holm and
0/30 global-omnibus rejections. Artificial forecasts $\hat Y=\mu^F+\delta$ for
$\delta\in\{0,.1,.25,.5\}$ recover $\MSEmu=\delta^2$ over 800 seed--origin rows with maximum absolute
error $6.94\times10^{-18}$; Eq.~\ref{eq:oracle-risk} closes to $1.39\times10^{-17}$. These checks calibrate
the strict null and evaluator arithmetic.  Two further 100-seed controls pass through data generation,
model fitting, forecasting, and evaluation: increasing only innovation scale assigns 99.80\% of the
expected-risk increase to $\Delta B$, whereas deleting the last AR input leaves $\Delta B=0$ and produces
a positive $\Delta D=.000315$ with a seed-level 95\% interval $[.000262,.000372]$. Because total
degradation equals $\Delta D$ in this control, its added-distance fraction is $f=100\%$.
\IfFileExists{generated/stage4/positive_controls_table.tex}{\begin{table}[t]
\caption{\textbf{End-to-end evaluator calibration.} The pass rules were locked before execution. Positive-control intervals and the fraction bound use 50,000 paired seed-bootstrap draws.}
\label{tab:stage4-controls}
\centering
\footnotesize
\setlength{\tabcolsep}{3pt}
\begin{tabularx}{\linewidth}{>{\raggedright\arraybackslash}p{0.17\linewidth}>{\raggedright\arraybackslash}p{0.28\linewidth}>{\raggedright\arraybackslash}X>{\raggedleft\arraybackslash}p{0.07\linewidth}}
\toprule
Control & Prespecified pass rule & Observed attribution & Result \\
\midrule
Identical-data null & all paired effects are numerically zero & 400/400 profiles: $\Delta R=\Delta B=\Delta D=0$ & pass \\
Innovation $\sigma:.2\to.4$ & $\Delta B>0$ and $U_{.95}(f)<10\%$ & $\Delta B=0.120000$, $f=0.20\%$, $U_{.95}=0.22\%$ & pass \\
Delete last AR input & $\Delta B=0$ and $\Delta D>0$ & $\Delta B=0.000000$, $\Delta D=0.000315$, $f=100\%$, CI [0.000262, 0.000372] & pass \\
\bottomrule
\end{tabularx}
\end{table}
}{}
Table~\ref{tab:stage4-controls} records both the prospectively stated pass rule and the observed
attribution for each control. The null produces numerical zero through the full pipeline; the
innovation-scale and deleted-input controls assign end-to-end forecast changes to the environmental and
distance components in the intended directions.

\section{SMSE and deployment-weight sensitivity}
\label{app:smse-sensitivity}

The scale-relative metric is the mean of pointwise ratios
\begin{equation}
\SMSE_m^e(k,H)=\frac{1}{NH}\sum_{i=1}^N\sum_{h=1}^H
\left(\frac{Y^{e,(i)}_{t+h-1}-\hat Y^{e,(i)}_{m,t+h-1}}
{\max(\hat\sigma^{e,(i)}_{t,h},\epsilon)}\right)^2,
\qquad \epsilon=10^{-8}.
\label{eq:smse}
\end{equation}

SMSE therefore averages pointwise ratios rather than dividing aggregate mean squared error by one
aggregate scale. In outlier-variance feedback, the evaluator predictive scale can increase enough that
absolute errors grow while errors relative to the local origin-and-lead scale fall. This is distinct from
MASE, whose denominator is a global empirical training-sample naive-error scale. SMSE neither replaces
MASE nor constitutes probabilistic forecast calibration; it is a secondary local-scale diagnostic.

The minimum retained predictive scale is 0.168, so changing
$\epsilon$ from $10^{-8}$ through $0.05$ leaves every reported SMSE unchanged.  The lowest-scale 5\% of
points contribute 4.4--12.9\% of total SMSE, showing increased but not singular influence.  Changing from
the mean of pointwise ratios to the ratio of means changes paired effects by 0.038 on average and at most
0.176; increasing oracle paths from 2,048 to 8,192 changes an SMSE effect by at most 0.087.  Nevertheless,
model ranks are stable: Spearman correlation is 1.000 across all three path counts and at least 0.994
between mean-of-ratios and ratio-of-means.  We therefore retain SMSE as a scale-relative secondary view,
with raw risk and probabilistic calibration reported as separate estimands.

The TimesFM random-arm audit contains 540 matched rows: mean MSE is 5.402, median 0.187, 5\%-trimmed mean
1.591, maximum 183.9, and the top 1\% contributes 20.2\% of total MSE.  At $H=192$, its zero/adapted
quantile-crossing rates are 91.0\%/98.3\% when defined as ``any crossing anywhere in the path,''
5.73\%/15.48\% per individual lead, and 1.08\%/2.87\% per adjacent quantile pair.  The first definition
grows mechanically with horizon; the headline reliability statistic therefore uses the individual-lead
rate.

For a deployment weighting $w$, we report
$R_m(w)=\sum_{k,h}w(k,h)A_m(k,h)$ without discarding the underlying profile.  Across uniform, short-lead,
long-lead, trend-heavy, tail/event-heavy, and regime-heavy choices, pairwise rank correlations span
0.423--0.989 (median 0.890), showing why deployment weights should remain explicit whenever the profile
is aggregated.

\begin{figure}[tbp]
\centering
\includegraphics[width=0.97\textwidth]{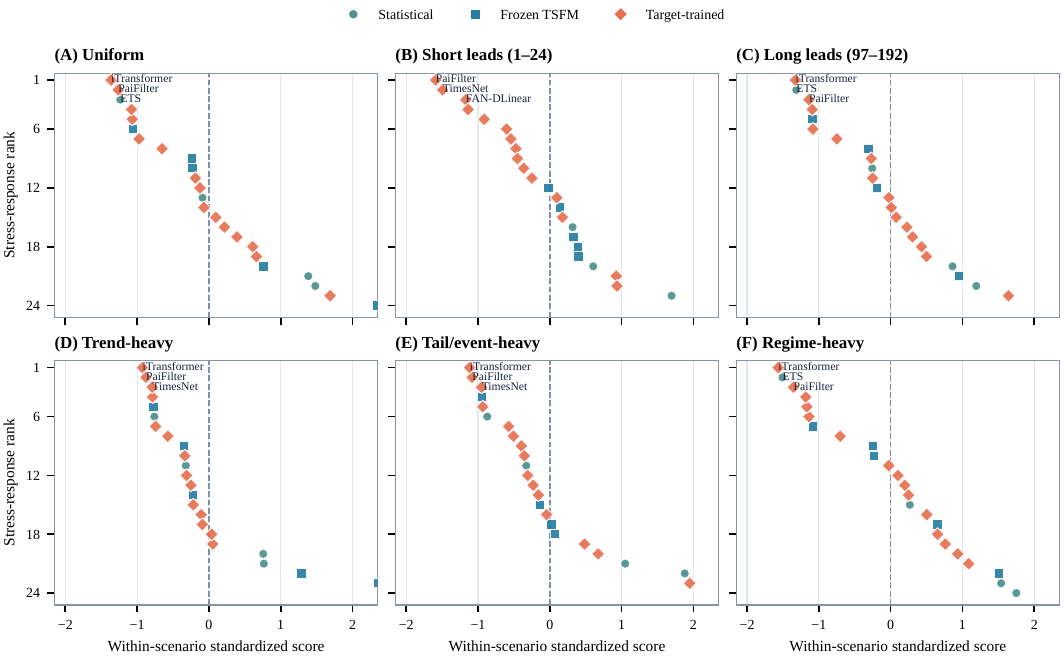}
\caption{\textbf{Deployment weighting changes stress-response rankings.} Each panel contains all 24
deployable forecasters; lower rank means a smaller weighted forecast--oracle-distance response. Scores
are standardized within each scenario, the dashed vertical line is that scenario's mean, and the top
three methods are labeled.}
\label{fig:deployment-rankings}
\end{figure}

Figure~\ref{fig:deployment-rankings} displays the corresponding model-level rankings rather than only
their correlations. Short-lead weighting promotes a different set of target-trained methods, while
uniform and long-lead weighting favor ETS and Chronos-2. These are rankings of weighted stress response,
not rankings of absolute forecasting accuracy.

\section{Global omnibus inference}
\label{app:omnibus}

For metric $q$, let $A^{(q)}_{s,p,m}(k,h)$ be the paired effect for DGP seed $s$, protocol $p$, model $m$,
mechanism $k$, and lead $h$.  The design-balanced statistic is
\begin{equation}
T_q=\sum_{p,m,k,h}w_{p,m,k,h}
\left[N_{\mathrm{seed}}^{-1}\sum_{s=1}^{N_{\mathrm{seed}}}A^{(q)}_{s,p,m}(k,h)\right]^2,
\end{equation}
where $w_{p,m,k,h}=(P K_p M_p H)^{-1}$.  Thus protocols have equal total weight and, within each protocol,
its available mechanisms, models, and leads are equally weighted.  The DGP seed is the independent unit.
Under sign exchangeability of the paired-difference vectors, each of the $2^{10}$ permutations flips the
complete 21,312-cell effect vector of a DGP seed, preserving all within-seed dependence.  The test is
finite-sample exact under sign exchangeability, which is stronger than independence across DGP seeds.
Primary $p$-values are Holm-corrected across raw MSE, forecast--oracle distance,
and SMSE.  Rejecting the joint-zero null
detects a systematic response somewhere in the evaluated grid. Cell-level and held-out analyses provide
the corresponding model, mechanism, and lead localization.

\section{Exact data-generating settings}
\label{app:dgp}

\paragraph{Initialization and timing.}
Every univariate case has length 4,096 and is generated by
\texttt{numpy.random.default\_rng(dgp\_seed)} with seeds 1101--1110.  There is no discarded burn-in and no
within-series change point: \Base{} and \Shift{} are separate full paths whose stated parameters apply
from index zero.  Forecast origins are 3,712 and 3,904 (stride 192), with $C=512$.  The common 38-cell
comparison is exactly the following: trend uses Base $b=0$ versus strong $b=.002$; innovation shape uses
Gaussian Base versus standardized Student-$t_3$; GARCH persistence and Markov switching use their Base
and Shift roles.  These eight cases are evaluated at $H\in\{1,24,96,192\}$ (32 cells).  Outlier-variance feedback
uses its Base and Shift at $H\in\{24,96,192\}$ (six cells).  Moderate trend, skew/mixture innovation, and
all other manifest roles are outside this exact cross-protocol intersection.  Oracle paths use $S=2,048$
and RNG seed
\begin{equation}
s_{\rm oracle}=(1000003s_{\rm DGP}+10007t+101H+2026083101)\bmod 2^{32}.
\end{equation}

\paragraph{Equations.}
Let $z_t\sim\mathcal N(0,1)$ and $B_t\sim\mathrm{Bernoulli}(p)$ with independent symmetric sign $S_t$.
The implemented families are
\begin{align}
Y_t&=a+bt+\sigma z_t,\\
Y_t&=A\sin\!\left(2\pi\sum_{j\le t}f_j\right)+\sigma z_t,\\
Y_t&=a+bt+\sigma Z_t,\\
Y_t&=a+bt+\epsilon_t,\quad
\epsilon_t=\sqrt{h_t}z_t,\quad
h_t=\omega+\alpha\epsilon_{t-1}^2+\beta h_{t-1},\\
Y_t&=\mu_{Q_t}+\sigma_{Q_t}z_t,\quad Q_t\mid Q_{t-1}\sim P,\\
Y_t&=a+bt+\sigma z_t+B_tS_tA,\\
L_t&=\rho L_{t-1}+B_tS_tA,\quad Y_t=a+bt+L_t+\sigma z_t,\\
Y_t&=a+bt+\sqrt{h_t}z_t+B_tS_tA,\\
h_{t+1}&=\sigma^2+\rho_v(h_t-\sigma^2)\notag\\
&\quad+\gamma(B_tS_tA)^2.
\end{align}
For innovations, $Z_t$ is centered and variance-standardized: Student-$t_\nu$ is multiplied by
$\sqrt{(\nu-2)/\nu}$; skew-normal is centered and divided by its analytic standard deviation; and the
two-Gaussian mixture is centered and divided by
$\sqrt{\sigma_c^2+4w(1-w)m_c^2}$.  GARCH starts at
$h_0=\omega/(1-\alpha-\beta)$.  The two-state Markov chain starts in its stationary distribution; with
equal off-diagonal transition probability this is $(1/2,1/2)$.

\begin{table*}[!htbp]
\caption{Univariate mechanism pairs and parameter settings.  Shared defaults are $a=.5$, $b=.002$, and
$\sigma=.2$ unless replaced below; parameters not named within a pair are identical.}
\label{tab:dgp-cases}
\centering
\scriptsize
\setlength{\tabcolsep}{2.8pt}
\resizebox{\textwidth}{!}{%
\begin{tabular}{lllll}
\toprule
Pair & Case role(s) & Mechanism & Exact parameters & Formal horizons \\
\midrule
Trend strength & Base; moderate; strong & linear mean & $b=0;.001;.002$, $a=.5,\sigma=.2$ & 1,24,96,192 \\
Frequency acceleration & Base; Shift & sinusoidal mean & $A=1$, $f:1/48\!\to\!1/48;1/48\!\to\!1/12$, $\sigma=.2$ & 1,24,96,192 \\
Innovation shape & Base; $t_3$; skew; mixture & standardized $Z_t$ & Gaussian; $\nu=3$; shape $=5$; $w=.5,m_c=2,\sigma_c=1$ & 1,24,96,192 \\
GARCH persistence & Base; Shift & GARCH(1,1) & $(\omega,\alpha,\beta)=(.0272,.08,.75);(.0048,.12,.85)$ & 1,24,96,192 \\
Markov rate & Base; Shift & two-state Gaussian & $\mu=(-.5,1.5)$, $\sigma=(.2,.2)$, stay $=.97;.75$ & 1,24,96,192 \\
Additive outliers & Base; strong & symmetric events & $(p,A)=(0,0);(.02,2)$ & 1,24,96,192 \\
Level recovery & Base; slow; persistent & level events & $(p,A,\rho)=(0,0,.5);(.05,.75,.9);(.05,.75,1)$ & 24,96,192 \\
Outlier-variance feedback & Base; Shift & event-driven scale feedback & $(p,A)=(.05,1.5)$; Shift adds $(\rho_v,\gamma)=(.85,.2)$ & 24,96,192 \\
\bottomrule
\end{tabular}
}
\end{table*}

Table~\ref{tab:dgp-cases} makes the mechanism library auditable at parameter level. The common discovery
intersection uses the named Base/Shift roles only; moderate trend, skewed/mixture innovations, and the
remaining event variants are documented extensions excluded from that intersection by construction.

\paragraph{Oracle availability.}
The deterministic/innovation cases use their known mean; GARCH and outlier-variance-feedback means are $a+b(t+h-1)$;
the Markov mean is $\pi_tP^{h-1}\boldsymbol\mu$ after filtering observations strictly before the origin;
and a pre-origin level state contributes $\rho^hL_{t-1}$. These means are analytic for every formal
univariate case.  One-step variances are analytic for every family; multi-step variance is analytic for
the homoskedastic and Markov cases, while the reported scale for GARCH and event paths is the $S=2,048$
Monte Carlo predictive-path variance. GARCH and Markov filtering use observed pre-origin values. The
persistent-level and outlier-feedback evaluators additionally use pre-origin event marks to reconstruct the
generator state, while the supplemental stochastic-volatility evaluator uses the pre-origin latent log
variance. These variables constitute the declared $Z_{<t}^{\mathrm{DGP}}$ in
$\mathcal F_t^{\mathrm{eval}}$. No future latent state, event, variance, or observation enters the oracle.

\paragraph{Multivariate relation shifts.}
For $d=4$, $\mathbf Y_t=A_t\mathbf Y_{t-1}+\boldsymbol\epsilon_t$ with
$\boldsymbol\epsilon_t\sim\mathcal N(0,.2^2I)$.  After a 512-step burn-in, 4,096 points are retained;
the change occurs at index 3,712, which is also the first origin, and the second origin is 3,904.
The diagonal-only reference is $.55I$.  The directed-ring reference has diagonal $.45$ and entries
$(1,2),(2,3),(3,4),(4,1)=.18$; the strength Shift changes those four entries to $.34$, and the topology
Shift changes them to $-.18$.  Every pre/post matrix is required to have spectral radius below one.
The analytic oracle recursively applies the appropriate pre/post matrix to the last observed vector;
all cases use seeds 1101--1110 and $H\in\{24,96,192\}$.

The held-out Stage-3A confirmation uses seeds 3201--3220, $H=192$, and origins 3,712/3,904.  In the
70/10/20 loader, the scaler and training observations end at index 2,866 and validation targets end at
3,276, both before the change.  \Base{} and \Shift{} are identical through index 3,711.  Four
target-trained models are launched separately with initialization 6101 but receive identical
pre-change fit data; two TSFMs reuse their checkpoints.  An output audit finds bitwise-identical
\Base/\Shift{} forecasts at origin 3,712 for every one of the 120 model--seed comparisons.  At origin
3,904, the 512-step context contains 192 post-change values but no weight update.  Thus the two origins
measure two distinct estimands: immediate OOD and passive contextual response.

\paragraph{Matched-difficulty topology audit.}
The topology audit keeps $d=4$, the diagonal coefficient .45, innovation variance .04, $C=512$, and
$H=192$ fixed. The Block system has two communities $\{1,2\}$ and $\{3,4\}$, with weak
cross-community edges .02; the Hub system uses channel 1 as the center and keeps weak return edges .02.
For $c_B\in\{.18,.305984\}$ and $c_H\in\{.18,.440642\}$,
\begin{equation}
A_{\rm Block}(c_B)=
\begin{bmatrix}
.45&c_B&.02&0\\ c_B&.45&0&.02\\ .02&0&.45&c_B\\0&.02&c_B&.45
\end{bmatrix},\qquad
A_{\rm Hub}(c_H)=
\begin{bmatrix}
.45&.02&.02&.02\\ c_H&.45&0&0\\ c_H&0&.45&0\\c_H&0&0&.45
\end{bmatrix}.
\end{equation}
The Shift spectral radii are .775984 and .612599. Coefficients were fixed analytically by propagating
$\Sigma_h=A\Sigma_{h-1}A^\top+.04I$ from $\Sigma_0=0$ and matching
\begin{equation}
\Delta B=\frac{1}{192\times4}\sum_{h=1}^{192}\operatorname{tr}
\left(\Sigma_h^{\Shift}-\Sigma_h^{\Base}\right)
\end{equation}
to the Ring $.18\to.34$ reference value .0123485220. The resulting Block and Hub values are
.0123485904 and .0123484937, with relative mismatch below $6\times10^{-6}$. Block and Hub use independent
seeds 4601--4620 and 4701--4720. Four target-trained models share one pre-change fit per model--seed;
Chronos-2 and TTM reuse fixed checkpoints. Table~\ref{tab:stage5-topology} gives every primary cell.

\begin{figure*}[t]
\centering
\includegraphics[width=\textwidth]{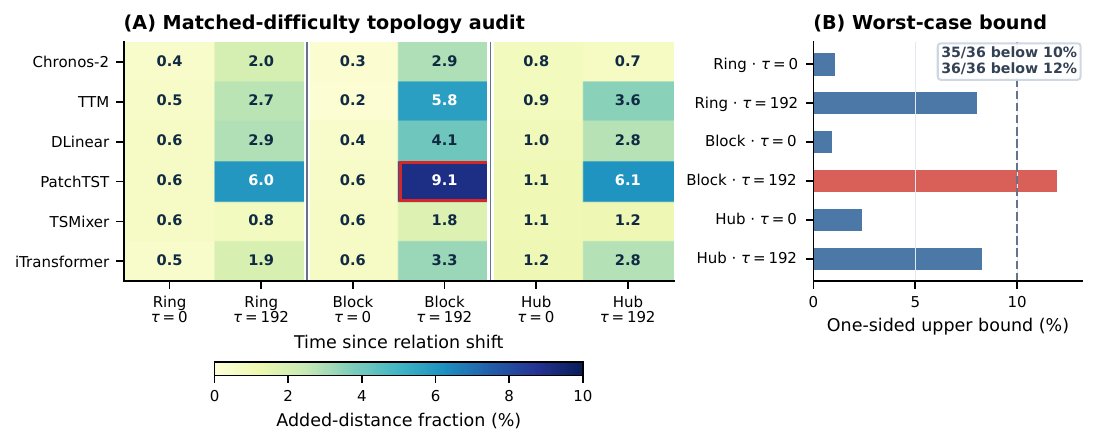}
\caption{\textbf{Environmental majority persists across matched-difficulty Ring, Block, and Hub
relations.} (A) Added-distance fractions for six models immediately after the relation shift
($\tau=0$) and after 192 post-shift observations ($\tau=192$); the systems match
$\Delta B\approx0.01235$. (B) The largest simultaneous one-sided 95\% upper confidence bound across
models for each topology--origin combination.}
\label{fig:topology-audit}
\end{figure*}

Figure~\ref{fig:topology-audit} visualizes the same cells listed in Table~\ref{tab:stage5-topology}.
Matching $\Delta B$ before model evaluation makes this a test of structural generalization rather than a
comparison between arbitrarily different environmental-risk changes.

\begin{figure*}[tbp]
\centering
\includegraphics[width=0.97\textwidth]{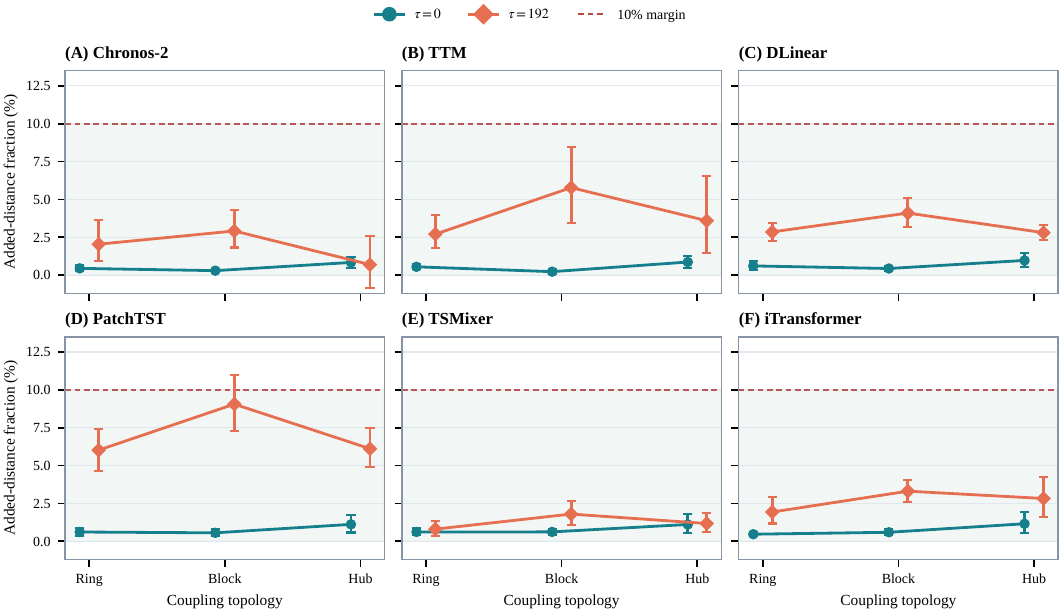}
\caption{\textbf{Model-resolved attribution across matched-difficulty coupling topologies.} Points show
the added-distance fraction at the immediate and later origins; error bars are two-sided 95\%
paired-seed bootstrap intervals. The green band lies below the prespecified 10\% margin; formal
simultaneous one-sided upper bounds are reported in Table~\ref{tab:stage5-topology}.}
\label{fig:topology-by-model}
\end{figure*}

Figure~\ref{fig:topology-by-model} expands the aggregate audit into six identically scaled model panels.
The immediate response remains small across topologies, while the later origin exposes model-specific
variation, including the larger Block response for PatchTST. This separates the shared environmental
majority from heterogeneous passive contextual responses.

\FloatBarrier

\section{Complete model roster and outputs}
\label{app:model-specs}

The target-trained implementations are TimeKAN, TimeMixer, PaiFilter, TexFilter, TimesNet, SegRNN,
TSMixer, DLinear, Non-stationary Transformer, iTransformer, PatchTST, Autoformer, Koopa, SAN-DLinear, and
FAN-DLinear~\citep{wu2023timesnet,lin2023segrnn,chen2023tsmixer,zeng2023dlinear,liu2022nonstationary,
liu2024itransformer,nie2023patchtst,wu2021autoformer,liu2023koopa,liu2023san,ye2024fan,wang2024timemixer}.
PaiFilter and TexFilter are two executable FilterNet variants, hence 15 implementations but
14 architecture labels after grouping that family.  The frozen models are Chronos-2, TimesFM~2.5,
Moirai~2.0-small, TTM-R2, and Time-MoE-200M
\citep{ansari2025chronos2,das2024timesfm,woo2024moirai,ekambaram2024ttm,shi2024timemoe}. Chronos, TimesFM, and Moirai provide quantiles in addition to
point predictions; TTM and Time-MoE are evaluated as point-only.  Missing probabilistic entries mean
``unsupported output.''

The broader forecasting landscape includes residual MLPs and efficient Transformers
\citep{oreshkin2020nbeats,zhou2021informer,zhou2022fedformer,kim2024cats,wang2024timexer}, generative and
cross-modal pretrained forecasters~\citep{gruver2023llmtime,goswami2024moment,liu2024timer,
liu2025sundial,liu2025moiraimoe,chen2025visionts}, and synthetic or in-context learning approaches
\citep{lu2025ictsp,taga2025timepfn,faw2025incontext}. These works motivate the heterogeneous roster; the
evaluated coverage is the roster enumerated above. Online correction constitutes a distinct deployment
protocol from the fixed-checkpoint and matched-support arms used here~\citep{lee2025elf}.

\paragraph{Statistical and target-trained specifications.}
The five anchors are last-value naive, seasonal naive (period 24), AutoReg (maximum lag 12, constant and
trend), ETS (additive damped trend, no seasonality), and the DGP oracle mean (reference only).  All 15
target-trained models use the 70/10/20 chronological split, MSE, at most 30 epochs, patience 10,
training-only scaling, and validation-best selection; checkpoint selection uses validation data only.
Defaults inherited from \texttt{run.py} are Adam with learning rate $.001$, batch 32, $d=512$,
$d_{ff}=2048$, eight heads, two encoder layers, and dropout $.1$.  Table~\ref{tab:target-specs} lists every
active override. Parser arguments not consumed by an implementation are explicitly marked inactive.

\begin{table*}[!htbp]
\caption{Implementation settings for target-trained models.  ``lr/bs'' denotes learning rate and batch
size; $d$ and $f$ denote model and feed-forward width.}
\label{tab:target-specs}
\centering
\scriptsize
\setlength{\tabcolsep}{2.5pt}
\resizebox{\textwidth}{!}{%
\begin{tabular}{llll}
\toprule
Implementation & lr / bs & Main dimensions & Other fixed settings \\
\midrule
TimeKAN & $.01/128$ & $d=16,f=32,L=2$ & downsample layers/window $=2/2$, order 0 \\
TimeMixer & $.01/128$ & $d=16,f=32,L=2$ & average downsample layers/window $=2/2$ \\
PaiFilter & $.01/32$ & hidden 256 & label length 0 \\
TexFilter & $.001/32$ & embed/hidden $=512/512$ & dropout 0 \\
TimesNet & $.001/32$ & $d=512,f=2048,L=2$ & top-$k=5$, factor 3 \\
SegRNN & $.0001/32$ & $d=512$ & dropout .5; segment $1$ for $H=1$, else $2$ \\
TSMixer & $.001/32$ & two residual mixer blocks; $d=512$ & decoder-layer/factor arguments inactive \\
DLinear & $.01/32$ & seasonal/trend linear heads & moving average 25; encoder/decoder-layer/factor arguments inactive \\
Non-stationary Transformer & $.0001/32$ & $d=512$, 8 heads, $L=2/1$ & label 48; projector $[128,128]$ \\
iTransformer & $.0005/32$ & $d=f=512$, 8 heads, $L=3$ & dropout .1, factor 3 \\
PatchTST & $.0001/32$ & three encoder layers, 4 heads & patch 16, factor 3 \\
Autoformer & $.001/32$ & two encoder/one decoder layers & label 48, moving average 25 \\
Koopa & $.0001/32$ & $d=512$, two encoder layers & factor 3 \\
SAN-DLinear & $.01/32$ & DLinear backbone & period 8; station lr $.0001$; pretrain 5 epochs \\
FAN-DLinear & $.01/32$ & DLinear backbone & top-$k=4$; auxiliary weight 1 \\
\bottomrule
\end{tabular}}
\end{table*}

Table~\ref{tab:target-specs} states every non-default implementation override, making clear that the
comparison uses executable procedures rather than architecture names alone. Shared settings are stated in
the surrounding text so the table can focus on model-specific choices.

The target-trained runner is this repository's modified Time-Series-Library-style \texttt{run.py}, executed
at local commit \texttt{d73a531b5c8b36b00a6920f5b10a7d4aacdb0296}. It adds explicit Python, NumPy,
and PyTorch seed control, generated-series loading, configurable artifact destinations, and a formal runner
that verifies each saved test target against the requested DGP slice. The repository metadata does not
retain an upstream commit identifier, so reproducibility is tied to the released local source rather than
claimed for a stock upstream checkout.

All use $C=512$ and direct $H\in\{1,24,96,192\}$ heads in the common grid; training seeds are
3101--3103.  The scaler is fit only on the training split and inverted before scoring.

\begin{table*}[!htbp]
\caption{Frozen TSFM checkpoints and forecasting settings.  Revision entries are 12-character prefixes of
immutable repository commits; the released machine-readable manifests retain every full hash.}
\label{tab:frozen-specs}
\centering
\small
\setlength{\tabcolsep}{3pt}
\begin{tabular}{p{0.13\textwidth}p{0.29\textwidth}p{0.17\textwidth}p{0.32\textwidth}}
\toprule
Model & Repository & Revision prefix & Context/horizon and output \\
\midrule
Chronos-2 & \path{amazon/chronos-2} & \texttt{29ec3766d36d} & $C=512,H\le192$; point + 9 quantiles \\
TimesFM 2.5 & \path{google/timesfm-2.5-200m-pytorch} & \texttt{1d952420fba8} & $C=512,H\le192$; point + 9 quantiles \\
Moirai 2.0 R-small & \path{Salesforce/moirai-2.0-R-small} & \texttt{30f43ff08c84} & $C=512,H\le192$; point + 9 quantiles \\
TTM-R2 & \path{ibm-granite/granite-timeseries-ttm-r2} & \texttt{d6a79570cac0} / \texttt{25f4a00a25e1} & $H=1,24,96$: native 512--96; $H=192$: native 512--192; point only \\
Time-MoE-200M & \path{Maple728/TimeMoE-200M} & \texttt{794591bfeb12} & $C=512,H\le192$ autoregressive; point only \\
\bottomrule
\end{tabular}
\end{table*}

Table~\ref{tab:frozen-specs} identifies the immutable checkpoint revision and output contract for every
frozen TSFM. Revision hashes prevent a repository alias from silently resolving to different weights;
the point/quantile distinction explains structural missingness in probabilistic metrics.

\paragraph{Matched adaptation.}
Every arm uses 64 deterministic pre-cutoff support windows and 64 later disjoint validation windows,
three training seeds (4101--4103), $C=512$, $H\in\{1,192\}$, and identical test origins.  Random arms
use the same architecture without pretrained tensors and select the validation-best epoch.  TTM trains
all parameters in the random arm and decoder/head only in the pretrained arm (AdamW, lr $.001$, batch 32,
20 epochs, patience 3).  TimesFM, Chronos, Moirai, and Time-MoE use rank-4 LoRA
($\alpha=8$, dropout .05)~\citep{hu2022lora}:
all linear modules for TimesFM; attention Q/K/V/O and output patch embedding for Chronos; attention and
FFN projections for Moirai; and attention plus expert gate/up/down projections for Time-MoE.  Their LoRA
learning rates are $10^{-4},10^{-5},10^{-4},10^{-4}$, respectively.  TimesFM uses 10 epochs/batch 16;
Chronos 40 steps/batch 16 with evaluation every 10 steps; Moirai 20 epochs/batch 16; and Time-MoE 10
epochs/effective batch 16 with gradient checkpointing.  TimesFM $H=192$ uses teacher-forced 128+64
training chunks and identical autoregressive evaluation; Moirai uses three 64-step chunks; Time-MoE uses
its native autoregressive heads. Probabilistic summaries use the three models with native quantile
outputs.
The TimesFM adaptation checkpoint is the Transformers-format revision
\texttt{5a9806b9b291fad9233b5249d88263f1846304d3}; all other adapted arms use the revisions in
Table~\ref{tab:frozen-specs}.

\begin{table*}[!htbp]
\caption{Matched adaptation results on the discovery grid.  Effects average origins
and three training seeds inside each of 10 DGP seeds.  Repair is
$(\text{zero}-\text{adapt})/\text{zero}$; intervals bootstrap DGP seeds and $p_H$ is from the paired
sign-flip test (finite-sample exact under sign exchangeability) with Holm correction across five models.
Repair values are computed from unrounded seed-level quantities rather than the displayed means.}
\label{tab:adapt}
\centering
\small
\resizebox{0.88\textwidth}{!}{%
\begin{tabular}{lrrrrr}
\toprule
Model & Zero & Adapt & $\Delta$ [95\% CI] & Repair [95\% CI] & $p_H$ \\
\midrule
TimesFM & .279 & \textbf{.253} & $-.026$ [$-.038,-.010$] & 9.3\% [3.9,13.5] & .031 \\
Chronos-2 & .271 & .271 & $-.0001$ [$-.0013,.0009$] & 0.04\% [$-0.3,0.5$] & .891 \\
Moirai & .300 & \textbf{.270} & $-.030$ [$-.042,-.017$] & 9.9\% [6.0,14.3] & .010 \\
TTM & .344 & \textbf{.256} & $-.088$ [$-.106,-.071$] & 25.5\% [21.5,30.1] & .010 \\
Time-MoE & .354 & \textbf{.255} & $-.098$ [$-.108,-.089$] & 27.8\% [25.2,31.1] & .010 \\
\bottomrule
\end{tabular}}
\end{table*}

Table~\ref{tab:adapt} compares zero-shot and matched-support adaptation within model and DGP seed. The
largest MSE repairs occur for TTM and Time-MoE, while the near-zero Chronos effect and the reliability
diagnostics below establish substantial model- and metric-level heterogeneity around the average effect.

\paragraph{Adaptation diagnostics beyond the aggregate table.}
Effects remain horizon-specific: Time-MoE random/zero/adapt MSE changes from 0.364/0.189/0.174 at $H=1$
to 0.307/0.485/0.321 at $H=192$.  TimesFM's random mean of 5.402 is unstable: its median is 0.187 and the
top 1\% of rows contribute 20.2\% of loss.  Accuracy and reliability can diverge: TimesFM adaptation lowers
MSE by 9.3\%, but $H=192$ coverage falls from .624 to .592 and individual-lead crossing rises from 5.73\%
to 15.48\% (adjacent-pair: 1.08\% to 2.87\%). The path-any rate of 91.0\%/98.3\% aggregates whether at
least one crossing occurs anywhere in a path; the individual-lead rate is therefore the headline
reliability statistic.

The complete machine-readable sources are
\texttt{configs/phase4f\_formal\_zero\_shot\_plan.json},
\texttt{phase4m\_formal\_target\_trained\_plan.json},
\texttt{phase5a\_formal\_statistical\_baselines.json}, and the phase5b/phase6 adaptation manifests;
checkpoint manifests additionally record weight hashes.

\paragraph{Software, hardware, and compute accounting.}
The released execution environment uses Python 3.11.15, PyTorch 2.5.1 with the CUDA 12.1 runtime, and
statsmodels 0.14.6. The formal target-trained grid ran on four NVIDIA RTX 4090 GPUs. Surviving rental
manifests do not record the host CPU and RAM, so we do not infer them retrospectively. Summing the elapsed
time stored for 28,350 target-training jobs gives approximately 625.8 aggregate GPU-job hours, including
433.9 hours for the core comparison, 114.9 for context sensitivity, and 77.0 for event recovery. These are
per-job elapsed sums and overlap under concurrent execution; they are not wall-clock duration. Formal
frozen-model invocations record about 1.08 aggregate process-hours, and statistical baselines about 0.22
CPU-hours. The formal oracle inventory contains 1,500 unique windows with 2,048 paths each, or 3,072,000
simulated paths and approximately 1.91 GiB of stored sample values; oracle CPU time was included in runner
wall times and was not logged separately. Postprocessing and compilation were performed on a node with
NVIDIA A800 80GB GPUs, 56 Intel Xeon Gold 6348 CPU cores, and approximately 1 TiB RAM. These totals are
computed from retained job and runner manifests rather than extrapolated from a pilot.

\FloatBarrier

\section{External-check scope and descriptive results}
\label{app:external-checks}

For the real-residual checks, robust STL is fitted to each complete donor series using period 24 for UCI
Air Quality and Metro Traffic and period 96 for UCI Electricity.  Centered, standardized contiguous
residual blocks are injected at scale $.2$ around the same linear center as the Gaussian reference.
Air Quality uses the hourly \texttt{C6H6(GT)} channel, treats $-200$ as missing, and interpolates
chronologically with endpoint extension only when required. Electricity uses the complete 15-minute
\texttt{MT\_250} meter without aggregation. Metro uses \texttt{traffic\_volume}, averages duplicate
timestamps, retains observations from 2015-09-30 to avoid the archive's multi-month gap, reindexes hourly,
and time-interpolates missing hours. The implementation calls statsmodels 0.14.6
\texttt{STL(..., robust=True)} with seasonal periods 24 or 96, seasonal smoother 7, the default smallest
admissible odd trend and low-pass windows, degree one, and jump one.
Air Quality and Electricity use ten DGP seeds; Traffic uses six because non-overlapping source blocks are
required.  Five frozen checkpoints are evaluated at $H\in\{1,24,96,192\}$ and two origins.  We retain
raw MSE and distance to the injected center. Because serially dependent empirical blocks lack a validated
conditional resampling law, the latter is reported as center distance and the results form a donor-level
descriptive sensitivity audit. Table~\ref{tab:external-residual} pools the dependent horizons and origins
within each DGP seed.

\begin{table}[H]
\caption{\textbf{Sensitivity to semisynthetic residual noise.} Each donor residual pool is extracted by robust STL on the named real series, centered and standardized, then injected at scale $.2$ around the same linear center as the Gaussian reference. Values pool four horizons and two origins within each DGP seed, preserving the seed as the unit of description. The reported center distance reflects the absence of a validated conditional residual resampler.}
\label{tab:external-residual}
\centering
\scriptsize
\setlength{\tabcolsep}{4pt}
\resizebox{\textwidth}{!}{%
\begin{tabular}{llrrrrr}
\toprule
Donor (DGP seeds) & Frozen model & Reference MSE & Donor MSE & $\Delta$ raw MSE & $\Delta$ center distance & Paired windows \\
\midrule
UCI Air Quality (10) & Chronos-2 & 0.0405 & 0.0369 & -0.0036 & +0.0055 & 80 \\
 & TimesFM & 0.0423 & 0.0764 & +0.0341 & +0.0389 & 80 \\
 & Moirai & 0.0505 & 0.0612 & +0.0108 & +0.0174 & 80 \\
 & TTM & 0.1169 & 0.1040 & -0.0129 & -0.0081 & 80 \\
 & Time-MoE & 0.1230 & 0.1384 & +0.0154 & +0.0211 & 80 \\
\midrule
UCI Electricity (10) & Chronos-2 & 0.0405 & 0.0378 & -0.0027 & +0.0078 & 80 \\
 & TimesFM & 0.0423 & 0.0497 & +0.0073 & +0.0198 & 80 \\
 & Moirai & 0.0505 & 0.0494 & -0.0011 & +0.0112 & 80 \\
 & TTM & 0.1169 & 0.0971 & -0.0199 & -0.0062 & 80 \\
 & Time-MoE & 0.1230 & 0.1437 & +0.0208 & +0.0302 & 80 \\
\midrule
UCI Metro Traffic (6) & Chronos-2 & 0.0441 & 0.0076 & -0.0366 & +0.0458 & 48 \\
 & TimesFM & 0.0466 & 0.0335 & -0.0131 & +0.0495 & 48 \\
 & Moirai & 0.0540 & 0.0548 & +0.0009 & +0.0426 & 48 \\
 & TTM & 0.1204 & 0.1045 & -0.0159 & +0.0145 & 48 \\
 & Time-MoE & 0.1263 & 0.0953 & -0.0310 & +0.0196 & 48 \\
\bottomrule
\end{tabular}}
\end{table}

The earlier FMV experiment crosses four frequency, four mean, and four volatility mechanisms over 64
three-way datasets, alongside 12 one-factor and 48 two-factor datasets, at four horizons.  It averages 12
implementation labels under one DGP/training seed, making it a hypothesis-generating factorial analysis.
The V-bearing aggregate below makes the cited interaction pattern auditable; formal claims rely on the
independent-seed experiments reported above.

\begin{table}[H]
\caption{\textbf{Exploratory FMV factorial results.} Results average 12 implementation labels under one DGP/training seed and serve as a hypothesis-generating factorial analysis. $\mathrm{MSE}_{true}$ is the deterministic predecessor of oracle distance and DED is the fraction outside the known scale band.}
\label{tab:fmv-exploratory}
\centering
\scriptsize
\setlength{\tabcolsep}{3pt}
\begin{tabular}{r rr rr rr}
\toprule
& \multicolumn{2}{c}{V only} & \multicolumn{2}{c}{F/M + V} & \multicolumn{2}{c}{F + M + V} \\
$H$ & $\mathrm{MSE}_{true}$ & DED & $\mathrm{MSE}_{true}$ & DED & $\mathrm{MSE}_{true}$ & DED \\
\midrule
24 & 0.0196 & 0.0073 & 0.3584 & 0.2699 & 0.4709 & 0.5390 \\
48 & 0.0204 & 0.0078 & 0.3340 & 0.2913 & 0.4587 & 0.5749 \\
96 & 0.0179 & 0.0069 & 0.3385 & 0.3196 & 0.4523 & 0.6143 \\
192 & 0.0170 & 0.0065 & 0.3484 & 0.3499 & 0.4494 & 0.6516 \\
\bottomrule
\end{tabular}
\end{table}

\FloatBarrier

\section{Detailed limitations}
\label{app:limitations}

The controlled oracle provides exact attribution within the specified generators. Semisynthetic residual
checks broaden the noise sources and report center distance because an exact conditional oracle is
unavailable for the empirical block resampler. The formal inferential unit is
10 DGP seeds in the discovery grid, followed by three independent ten-seed held-out batches for the five
locked contrasts and separate 20-seed multivariate and finite-fourth-moment checks. The paired sign-flip
tests are finite-sample exact under sign exchangeability of the paired-difference vectors, an assumption
stronger than independence across DGP seeds. Dependent origins and training seeds are averaged inside each
DGP seed. With ten discovery seeds, bootstrap bands characterize profile shape; the global omnibus tests
for any systematic grid-wide response, and the held-out contrasts provide prespecified localization.

The Student-$t_3$ innovation has finite variance but no fourth moment.  Consequently, squared error can
have infinite variance and ordinary bootstrap MSE intervals may be unstable.  We treat that case as an
extreme-stress descriptive probe.  Our standardized Student-$t_5$ check supplies a finite-fourth-moment
condition and uses oracle expected risk; all six model intervals span zero. Because the design targets
detection and has no equivalence margin, this result narrows the extreme-tail observation to the $t_3$
setting while leaving small $t_5$ effects unresolved. More generally, the exact decomposition is specific to
squared loss and the conditional-mean Bayes act; absolute, quantile, and distributional scores require
different oracle targets.

Severity coverage is most complete for coupling strength, with three post-change levels and replication
in an eight-channel modular VAR; the supported scope is therefore the evaluated linear-Gaussian systems.
The univariate target-trained protocol measures the response of an environment-specific fitting procedure,
whereas the multivariate relation protocol fixes the pre-change fit and measures change-point OOD plus
passive contextual response. Model coverage comprises five named TSFMs, one adaptation budget, three
real-residual donors, and two multivariate systems. The one-seed FMV factorial grid serves as
hypothesis-generating evidence. We will release anonymized code, configuration hashes, result manifests,
and the retained lead-time arrays needed to reproduce all reported analyses.

\end{document}